\documentclass[lettersize,journal]{IEEEtran}
\usepackage{amsmath,amsfonts,bm}
\usepackage{algorithmic}
\usepackage{algorithm}
\usepackage{array}
\usepackage[caption=false,font=normalsize,labelfont=sf,textfont=sf]{subfig}
\usepackage{textcomp}
\usepackage{stfloats}
\usepackage{url}
\usepackage{verbatim}
\usepackage{graphicx}
\usepackage{cite}

\usepackage{hyperref}
\hypersetup{%
  colorlinks=true,%
  linkcolor={red},
  citecolor={blue},
  urlcolor={magenta},
  bookmarksnumbered=true,%
  bookmarksopen=true}

\usepackage[flushleft]{threeparttable}
\usepackage{multirow}
\usepackage{gensymb}
\usepackage[caption=false]{subfig}
\usepackage[T1]{fontenc}

\newcommand{\gaussian}{\mathcal{N}}
\newcommand{\real}{\mathbb{R}}

\newcommand{\matD}{\mathbf{D}}
\newcommand{\matE}{\mathbf{E}}

\newcommand{\matH}{\mathbf{H}}

\newcommand{\matS}{\mathbf{S}}
\newcommand{\matU}{\mathbf{U}}
\newcommand{\matV}{\mathbf{V}}

\newcommand{\matX}{\mathbf{X}}
\newcommand{\matY}{\mathbf{Y}}

\newcommand{\matSigma}{\bm{\Sigma}}

\newcommand{\vecy}{\mathbf{y}}

\newcommand{\vecmu}{\bm{\mu}}

\usepackage[normalem]{ulem}
\useunder{\uline}{\ul}{}
\usepackage{siunitx}
\usepackage[utf8]{inputenc}
\usepackage{booktabs} 
\usepackage{caption}
\usepackage[table]{xcolor}
\usepackage{multirow}

\definecolor{better}{RGB}{100,149,237}
\definecolor{worse}{RGB}{220,20,60}

\newcommand{\downpct}[1]{\textcolor{better}{\scriptsize{$\downarrow$#1\%}}}
\newcommand{\uppct}[1]{\textcolor{worse}{\scriptsize{$\uparrow$#1\%}}}
\newcommand{\std}[1]{\textcolor{gray}{\scriptsize{(#1)}}}
\newcommand{\stdb}[1]{\textcolor{gray}{\scriptsize{(#1)}}}
\usepackage{booktabs}
\usepackage{tabularx}
\usepackage{ragged2e}
\usepackage{siunitx}
\usepackage{threeparttable}  
\usepackage[dvipsnames]{xcolor}

\newcolumntype{C}[1]{>{\centering\arraybackslash}m{#1}}

\begin{document}

\title{Eye Gaze-Informed and Context-Aware Pedestrian Trajectory Prediction in Shared Spaces with Automated Shuttles: \\ A Virtual Reality Study}

\author{Danya Li, Yan Feng$^{*}$, Rico Krueger$^{*}$
\thanks{Danya Li and Rico Krueger are with the Department of Technology, Management and Economics at the Technical University of Denmark.}
\thanks{Yan Feng is with the Department of Transport \& Planning, Civil Engineering Geosciences at Delft University of Technology.}
\thanks{$^{*}$ shared last authorship.}
}



\maketitle

\begin{abstract}

Predicting pedestrian behavior around automated shuttles in shared spaces is critical for their safe deployment, yet existing approaches overlook a rich source of human-perspective information: fine-grained eye gaze dynamics. 
To address this gap, we conduct a Virtual Reality experiment in which pedestrians interact with automated shuttles under varying approach angles (45\textdegree, 90\textdegree, 135\textdegree) and continuous-traffic conditions (single shuttle, two shuttles with 3 or 5-second gaps), collecting synchronized motion, eye gaze, and head orientation data. To investigate to what extent, under what conditions, and in what form fine-grained eye gaze is informative for pedestrian motion prediction, we develop a multi-modal prediction model that fuses these signals through modality-specific encoders, and systematically ablate gaze representations against head orientation and situational context. We report three main results. First, the predictive value of eye gaze is angle-dependent and tightly coupled with eye-head-body coordination: at acute angles where pedestrians actively redirect gaze to acquire the shuttle, eye gaze carries information that head orientation alone misses. Second, continuous gaze orientation outperforms categorical semantic fixation labels, with the optimal encoding frame (global or body-relative) depending on whether gaze is used alone or jointly with context. Third, eye gaze and situational context provide complementary predictive information: their combination reduces final displacement error (FDE) by 8.47\%, close to the sum of their individual contributions. Together, these findings highlight the value of incorporating human perceptual signals into pedestrian behavior prediction and motivate a human-centered complement to vehicle-centric modeling approaches. Our code are available at \url{https://github.com/danyayay/GazeX.git}. 
\end{abstract}


\section{Introduction}

Accurately predicting pedestrian trajectories is essential for the safe and efficient operation of automated vehicles in urban environments. 
Prior work in cognitive science and neuroscience has shown that eye gaze is not merely a passive reflection of visual attention but an active component of the perception-action loop: they are tightly coupled with subsequent actions in everyday behavior~\cite{land2009looking, patla2003far}. Moreover, fine-grained gaze dynamics reveal where pedestrians direct attention in real time, and thus may provide information about pedestrian intent before it becomes observable in body motion alone.

Despite this potential, the integration of fine-grained continuous eye-tracking signals into trajectory prediction models remains underexplored~\cite{hasan_mx-lstm_2018, rasouli_are_2017}. Existing approaches primarily rely on motion data and map-based context, sometimes augmented with skeleton, appearance, or individual features~\cite{zhang_pedestrian_2023}, but rarely incorporate attention explicitly. When attention is considered, it is typically approximated by head orientation due to the lack of eye-tracking data, and this proxy has been shown to improve prediction performance~\cite{hasan_mx-lstm_2018, ridel_understanding_2019, herman_pedestrian_2022, mo2022multi}. 
A parallel line of work uses vehicle-mounted cameras to estimate coarse semantic cues such as eye contact or target-directed attention~\cite{rasouli_are_2017, belkada_pedestrians_2021, murakami_pedestrians_2024}.
Meanwhile, studies in pedestrian--AV interaction have collected richer eye-tracking data, but these are typically aggregated for human factor analysis \cite{bindschadel_active_2022, bindschadel_two-step_2022, lanzer_interaction_2023} rather than used as predictive inputs for motion models. Together, these limitations suggest that the predictive value of fine-grained gaze dynamics remains insufficiently understood.

A second gap concerns the role of interaction geometry. Research in psychology shows that eye, head, body form a coordinated system, activated in a sequenced and context-dependent manner \cite{freedman2008coordination, franchak2021adapting, saeb2011learning}. Critically, the coordination pattern varies depending on the angular relationship between a person's current orientation and a target \cite{luo_eye-head-body_2022}. For small angular offsets, gaze shifts are primarily achieved through eye movements, whereas larger offsets progressively activate head and body rotations \cite{luo_eye-head-body_2022}. This suggests that the relationship between gaze and subsequent motion may vary across interaction geometries, potentially affecting the predictive value of gaze signals. Yet, most pedestrian--AV studies and datasets have predominantly focused on perpendicular crossings \cite{li2025analyzing}, leaving this geometric dependence largely unexplored.

The aim of this work is to investigate the predictive value of fine-grained eye gaze for pedestrian motion prediction in pedestrian--automated shuttle interactions. More specifically, we ask: (i) to what extent does fine-grained gaze improve prediction beyond motion and head orientation; (ii) under what conditions, particularly which approach geometries, is gaze most informative; and (iii) which gaze representation is most useful for prediction. We further examine how gaze interacts with situational context and whether the two carry overlapping or complementary information for human behavior prediction.

Answering these questions requires data that existing datasets do not provide, namely synchronized eye-tracking and trajectory data under controlled interaction geometries in pedestrian--shuttle shared spaces. To address this gap, we design and conduct a Virtual Reality (VR) experiment that systematically varies interaction angles between pedestrians and automated shuttles. The experiment takes place in a shared space, an unstructured environment without formal formal separation between modes and no strict traffic rules, where interactions can occur from multiple directions \cite{clarke2006shared}. This design choice serves a dual purpose: it enables us to evaluate the predictive value of fine-grained eye gaze across conditions eliciting different eye-head-body coordination patterns, and it captures the geometric diversity of shared-space encounters that prior work has largely overlooked \cite{li2025analyzing}. 

\begin{figure*}[t]
    \centering
    \includegraphics[width=\linewidth]{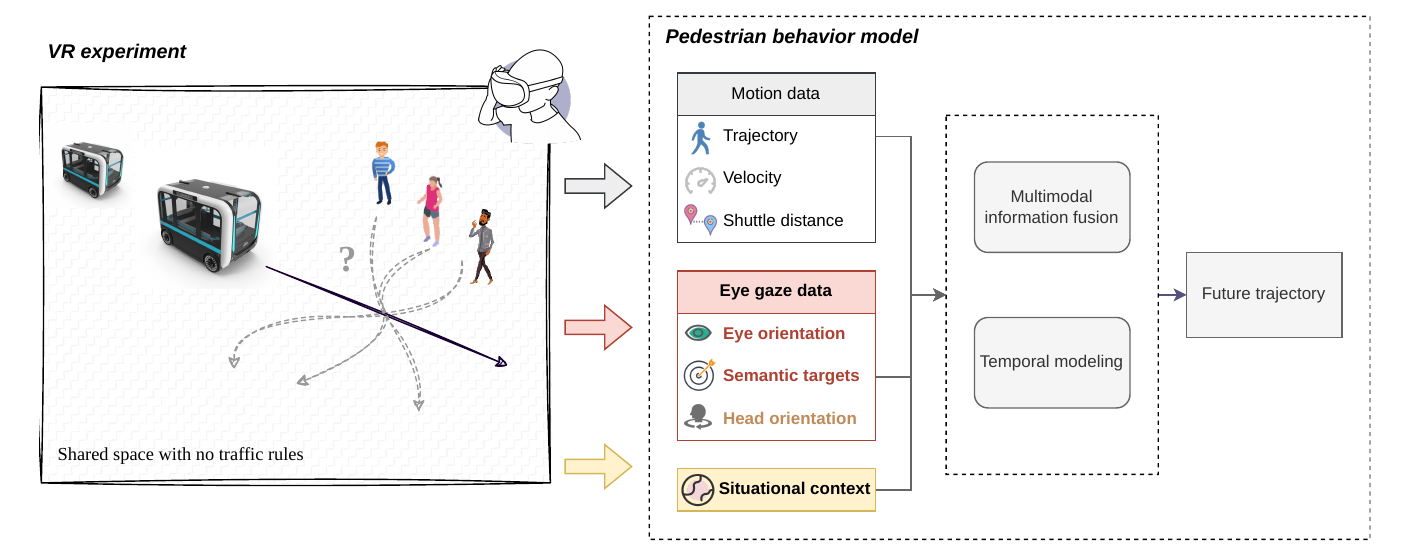}
    \caption{An overview of our paper.}
    \label{fig:overview}
\end{figure*}

We focus on automated shuttles because they commonly operate in low-speed shared environments where pedestrian interactions are frequent and differ from those involving on-road passenger vehicles \cite{woodman_gap_2019}. Reliable behavioral models in these settings are critical for shuttle navigation and for the broader integration of automated systems into shared urban spaces. 
To model pedestrian behavior, we develop a multi-modal prediction model fusing motion status, fine-grained eye gaze, and situational context through modality-specific long-short term memory (LSTM) encoders. 
Fig.~\ref{fig:overview} provides an overview of our approach.

Our contributions in this paper are threefold:
\begin{itemize}
\item We introduce a VR dataset that jointly varies two factors rarely studied together in pedestrian trajectory prediction: approach angle (45°, 90°, 135°) and continuous traffic (single shuttle, two shuttles with 3 or 5-second gaps). Using this dataset, we characterize how these factors affect hesitation, path deviation, gaze allocation, and proxemic preferences.
\item We systematically evaluate multiple representations of eye gaze to explore how much, under what conditions, and in what form they can be informative for pedestrian behavior prediction, including a comparison with head orientation as a common proxy. 
\item We quantify the contribution of situational context variables (approach angle, shuttle behavior, eHMI, continuous traffic) and show that gaze and context provide complementary rather than redundant information.
\end{itemize}



The rest of the paper is organized as follows: Sec.~\ref{sec:related_work} reviews related literature. Sec.~\ref{sec:methodology}, \ref{sec:experiment}, and \ref{sec:implementation} describe the modeling framework, VR experiment, and implementation details. Sec.~\ref{sec:results} and \ref{sec:discussion} present, analyze, and discuss the results with limitations and future directions. We conclude the paper in Sec.~\ref{sec:conclusion}. 

\section{Related works} \label{sec:related_work}

\subsection{Context-aware trajectory prediction}

Trajectory prediction has moved toward context-aware approaches that leverage information from multiple sources. 
Commonly used contexts include scene information represented by high-definition maps \cite{hu_collaborative_2020, kothari2023motion}, social context showing the influence of surrounding agents \cite{xiang2024socialcvae, xu_groupnet_2022}, and behavioral contexts encoding intentions and goals \cite{mangalam2021goals} or representing actions \cite{munir2025context}. These forms of context have shown strong predictive power. 

A smaller number of studies have examined situational context variables that capture external or scenario-level factors shaping the interactions \cite{kalatian_context-aware_2022, zhang2023cross, zhang2020research}. 
For instance, \cite{kalatian_context-aware_2022} examined the effects of road specifications (lane width, type of road), traffic parameters (speed limit, arrival rate), and environmental conditions (weather conditions, time of the day) while \cite{zhang2023cross} modeled the effects of crossing facilities. A number of studies have also investigated the effects of external human-machine interfaces (eHMIs) on pedestrian--AV interaction \cite{feng2023effect, feng_does_2024, faas2020external}. 
These situational factors have been more commonly investigated in the human factors field, but their quantitative effects for trajectory prediction are still limited. 



Our work explores the effects of several situational context factors, including approach geometry, vehicle behavior, eHMI presence, and continuous traffic, on the predictive power of pedestrian models for shared spaces.

\subsection{Pedestrian--AV interaction in shared spaces} \label{sec:related_work_shared_spaces}

Research on pedestrian--AV interaction has primarily focused on modeling crossing behavior in structured settings such as intersections and crosswalks~\cite{interactiondataset, ettinger2021large, caesar2020nuscenes, Rasouli2019PIE}. In these settings, interaction modeling is often guided by geometric structure and, in some cases, right-of-way assumptions embedded in structured road layouts. 

In contrast, modeling in unstructured shared spaces where no explicit traffic rules are imposed remains considerably less developed~\cite{golchoubian_pedestrian_2023}. The few available real-world datasets remain limited in scale~\cite{pascucci_discrete_2017, yang2019top, yang2019top, zhou2020developing}), and simulator-based studies have largely focused on perpendicular interactions \cite{li2025analyzing, feng_does_2024, woodman_gap_2019, andrijanto_application_2022}. However, shared spaces in practice involve a wide range of encounter geometries. Evidence from human-robot interaction suggest that human responses vary with approach angle \cite{walters2007robotic, garrell2013proactive}, but this has not been investigated for pedestrian--AV interactions \cite{predhumeau_pedestrian_2023}.

Moreover, existing shared-space studies have focused on conventional passenger vehicles or homogeneous pedestrian interactions, while automated shuttles---a natural agent in shared space environments---are often overlooked \cite{woodman_gap_2019}. Evidence suggests pedestrians perceive shuttles differently from conventional vehicles \cite{nunez_velasco_studying_2019} and adapt their behavior when interacting with small automation-capable vehicles \cite{zhou2020developing}. Field experiments with shuttles \cite{de_ceunynck_interact_2022} have been limited to basic statistical analysis due to the rarity of critical situations at low operating speeds.

From a methodological perspective, modeling approaches for shared-space interactions span expert-based, data-driven, and hybrid categories. Social force models (SFMs) \cite{helbing_social_1995} are the most common expert-based approach, capturing repulsive and attractive forces \cite{hossain2020conceptual, predhumeau2022agent, rashid2024simulation}; \cite{anderson_off_2020} extended this with a risk-based attention mechanism for predicting yielding. Data-driven methods, particularly LSTM-based models, achieve high accuracy by integrating spatial and temporal features \cite{kampitakis2023shared, cheng2018modeling, fafoutellis2023deep, yang2023prediction}. Hybrid approaches seek to balance predictive accuracy with interpretability by combining both paradigms \cite{johora2020agent, yang2024hierarchical}.

\subsection{Eye gaze in pedestrian behavior analysis}

In the trajectory prediction literature, head orientation is often used as a proxy for human attention, as eye-tracking data is typically unavailable \cite{hasan_mx-lstm_2018, ridel_understanding_2019, herman_pedestrian_2022, mo2022multi}. Extracted from vehicle-perspective visual input \cite{kim_pedestrian_2020, ridel_understanding_2019, herman_pedestrian_2022}, head pose has been shown to correlate with movement direction at high speeds \cite{hasan_mx-lstm_2018} and to improve trajectory prediction when combined with motion data \cite{ridel_understanding_2019}. These findings highlight head orientation as a useful attention indicator, but whether fine-grained eye gaze data can further enhance prediction remains unexplored.

Eye-tracking data has been applied in several VR-based studies on pedestrian--AV interactions in AV interface evaluation. 
Analysis typically relies on Areas-of-Interest (AoI), aggregating visual attention into regions ranging from general traffic scenes \cite{lanzer_interaction_2023, feng_does_2024} to specific vehicle components \cite{bindschadel_two-step_2022}.
These aggregated metrics, such as fixation duration, frequency, and distribution, serve to evaluate AV interface design \cite{bindschadel_two-step_2022}, distraction effect \cite{lanzer_interaction_2023}, or approximate mental workload \cite{bindschadel_active_2022} rather than continuous gaze dynamics for predictive modeling.

A related line of work in virtual reality investigates the use of eye gaze for locomotion prediction in redirected walking scenarios \cite{bremer_predicting_2024, bremer2021predicting, stein2022eye, kim2024gaitway, jeon_f-rdw_2025}. These studies show that the predictive benefit of gaze is context-dependent: for example, gaze improves the prediction of speed adjustments in goal-directed walking \cite{bremer_predicting_2024, bremer2021predicting, stein2022eye}, while performance degrades in visually complex or high-distraction environments \cite{kim2024gaitway}. This suggests that the utility of gaze is not uniform, but varies with task demands and environmental conditions. However, whether similar context-dependent effects arise in pedestrian–AV interactions, particularly under varying interaction geometries and traffic conditions, remains an open question.

A more detailed summary is in Tab.~\ref{tab:related_work} of Appendix.~\ref{sec:appendix}.

\section{Methodology} \label{sec:methodology}

\subsection{Problem definition} \label{sec:problem_definition}



We aim to predict a pedestrian's future trajectory given their recent behavior and situational context. Let $\vecy_{nt} \in \real^2$ denote the 2D position of sample $n$ at time step $t$. Given an observation window of $T_p$ steps and a prediction horizon of $T_f$ steps, we seek to learn a mapping $\mathcal{F}$:
\begin{equation}
    \matY^{fut} = \mathcal{F}(\matY^{past};\; \matS,\; \matX)
\end{equation}
where $\matY^{past} \in \real^{N \times T_p \times 2}$ contains past trajectories, $\matY^{fut} \in \real^{N \times T_f \times 2}$ contains future trajectories, and $N$ is the number of samples. The model conditions on two additional input groups:
\begin{itemize}
    \item \textbf{Time-varying features} $\matS = \{\matV, \matD, \matE\}$: pedestrian speed $\matV \in \real^{N \times T_p \times N_v}$, distance to shuttles $\matD \in \real^{N \times T_p \times N_d}$, and eye gaze or head orientation data $\matE \in \real^{N \times T_p \times N_e}$.
    \item \textbf{Situational variables} $\matX \in \real^{N \times 4}$: static scenario-level factors including eHMI presence, shuttle yielding behavior, approach angle, and continuous traffic condition.
\end{itemize}

\subsection{Model architecture}
We develop a multi-modal trajectory prediction model that fuses the the fine-grained eye gaze signals, motion data, and contextual information through modality-specific encoders. We adopt a lightweight LSTM-based framework for two main reasons. First, LSTMs remain a well-established choice for modeling sequential behavior in shared-space interactions, as discussed in Sec.~\ref{sec:related_work_shared_spaces}. Second, our dataset size is modest relative to what is typically required to effectively train more data-intensive architectures such as Transformers.

\begin{figure}[t]
    \centering
    \includegraphics[width=\linewidth]{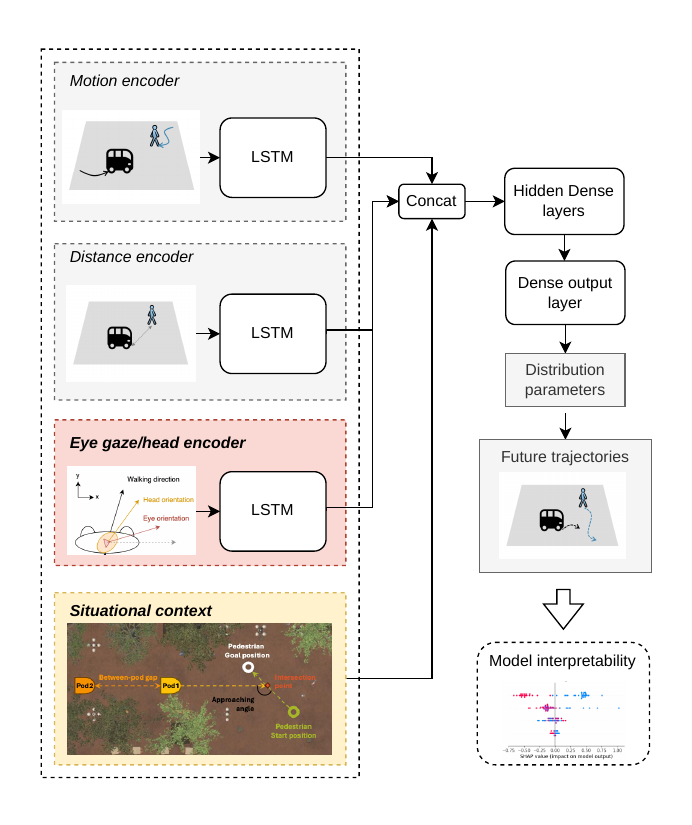}
    \caption{Model architecture.}
    \label{fig:model}
\end{figure}

The resulting model, referred to as \textit{GazeX-LSTM}, is illustrated in Fig.~\ref{fig:model} and consists of three stages: (i) encoding, where each modality is processed independently; (ii) fusion, where modality-specific representations are combined; and (iii) decoding, where future trajectories are predicted.

\paragraph{Encoding} Each group of time-varying features is processed by a dedicated LSTM encoder that summarizes the temporal dynamics into a fixed-length hidden representation:
\begin{align}
    \matH_m &= \text{LSTM}_m(\matY^{past}, \matV) \\
    \matH_d &= \text{LSTM}_d(\matD) \\
    \matH_e &= \text{LSTM}_e(\matE)
\end{align}
where $\matH_m$, $\matH_d$, and $\matH_e$ are the final hidden states of the motion, distance, and eye gaze encoders, respectively. Using separate encoders allows each modality to learn its own temporal patterns before fusion.
Past trajectory and velocity are grouped into a single motion encoder to avoid duplicate redundant temporal dynamics if separated.

\paragraph{Fusion and decoding} The encoded representations are concatenated with the static situational variables $\matX$ to form a joint representation, which is then passed through fully connected layers with ReLU activations:
\begin{align}
    \matH_h &= f_h([\matH_m,\; \matH_d,\; \matH_e,\; \matX]) \\
    [\vecmu,\; \matSigma] &= f_o(\matH_h)
\end{align}
The decoder outputs predictions for all $T_f$ future time steps in a single forward pass.

\paragraph{Probabilistic output} We model the future position at each time step as a bivariate Gaussian, $\vecy_{nt} \sim \gaussian(\vecmu, \matSigma)$, where $\vecmu \in \real^2$ and $\matSigma = \matU^T\matU \in \real^{2\times 2}$ with $\matU$ an upper triangular matrix. This Cholesky-style parameterization guarantees that the predicted covariance remains positive semi-definite during training without constrained optimization. The model thus predicts five distribution parameters ($\mu_x$, $\mu_y$, and three entries of $\matU$) per time step.

\subsection{Eye data representation}

\begin{figure*}[t]
    \centering
    \includegraphics[width=0.7\linewidth]{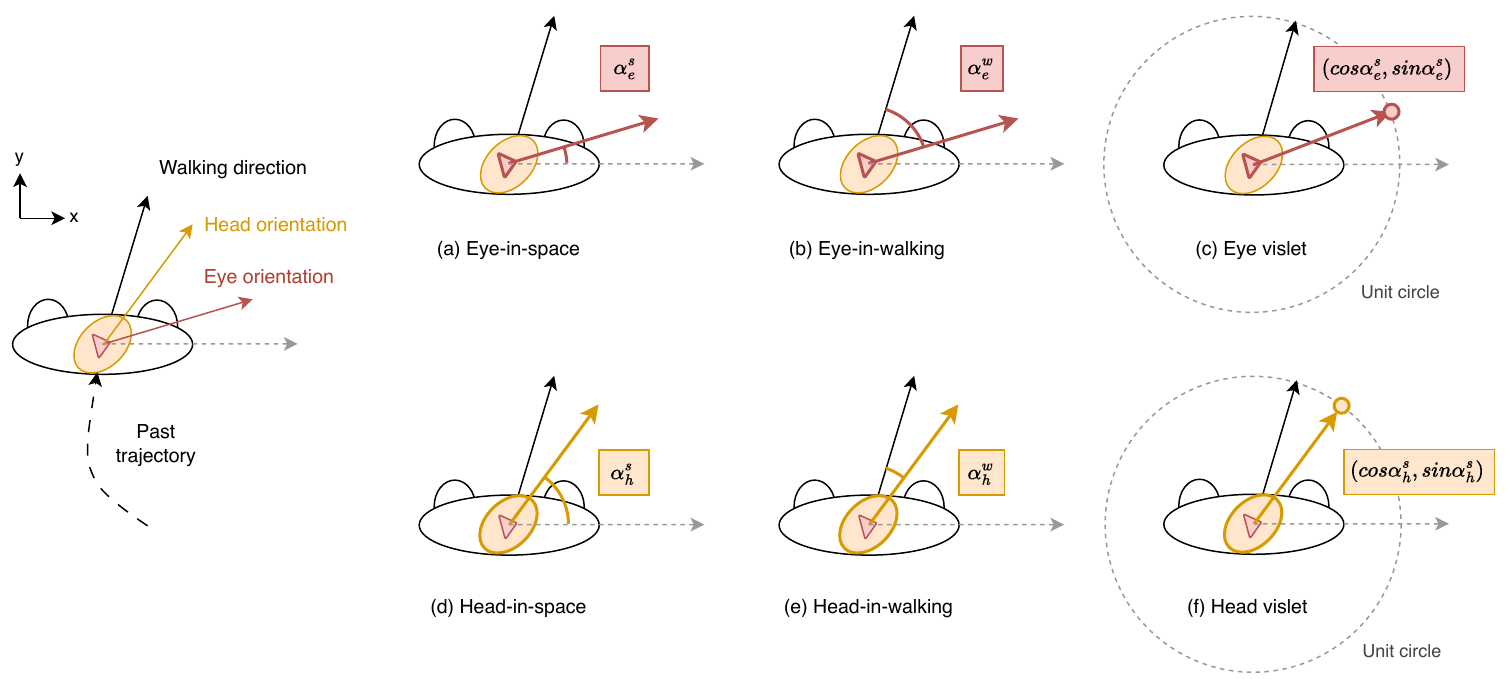}
    \caption{Representation of eye and head direction in our model. The first row shows the eye representation, and the second row shows the corresponding head representation. The first column uses the world frame as the reference axis, while the second column uses the walking direction as the reference axis. The third column shows the eye/head-in-space representations on a unit circle to avoid the discontinuity brought by angle representation.} 
    \label{fig:headeye_repr}
\end{figure*}

We systematically explore multiple representations of eye gaze and head orientation to evaluate how encoding choices affect prediction. All representations are summarized in Tab.~\ref{tab:eye_head_repr} and illustrated in Fig.~\ref{fig:headeye_repr}.

\paragraph{Eye orientation} We consider four representations of eye gaze direction, each capturing the horizontal (yaw) component from a top-down view (Fig.~\ref{fig:headeye_repr}a--c). These differ in their reference frame and encoding format. \textit{Eye-in-space} and \textit{eye-in-walking} encode the gaze angle in global and body-relative coordinates, respectively. The \textit{eye vislet} representation maps the gaze direction onto a unit circle, avoiding the discontinuity inherent in angular representations \cite{hasan_mx-lstm_2018}. 

\paragraph{Semantic gaze targets} Beyond continuous gaze direction, we also extract categorical labels indicating what the pedestrian is looking at. We define four levels of semantic granularity, ranging from coarse gaze event classification (fixation, saccade, noise) to fine-grained attention distribution across specific objects (leader shuttle, follower shuttle, goal, environment). These representations test whether knowing \textit{what} pedestrians attend to adds predictive value beyond knowing \textit{where} they look.

\paragraph{Head orientation.} In practice, eye gaze data is rarely available outside controlled experiments; head orientation extracted from vehicle-mounted cameras is the common substitute \cite{hasan_mx-lstm_2018, ridel_understanding_2019}. To quantify what is lost in this approximation, we derive three head orientation representations that mirror the eye orientation formats (Fig.~\ref{fig:headeye_repr}d--f): \textit{head-in-space}, \textit{head-in-walking}, and \textit{head vislet}.


\begin{table*}[t]
    \centering
    \caption{Summary of eye gaze and head orientation representations used in this study. ``Frame'' indicates whether the representation uses the world coordinate system (global) or the pedestrian's walking direction (local) as the reference axis. $N_e$ denotes the input dimensionality per time step.}
    \label{tab:eye_head_repr}
    \renewcommand{\arraystretch}{1.15}
    \begin{tabularx}{\textwidth}{
        >{\raggedright\arraybackslash}p{2cm}
        >{\raggedright\arraybackslash}p{3.4cm}
        >{\raggedright\arraybackslash}X
        c
        c
    }
        \toprule
        \textbf{Category} & \textbf{Representation} & \textbf{Description} & \textbf{Frame} & $N_e$ \\
        \midrule

        \multirow{4}{*}{\shortstack[l]{Eye\\orientation}}
        & Eye-in-space $\alpha^s_e$
        & Absolute eye direction angle in the world coordinate system.
        & Global & 1 \\
        & Eye-in-walking $\alpha^w_e$
        & Eye direction angle relative to the pedestrian's walking direction.
        & Local & 1 \\
        & Eye vislet $(\cos\alpha^s_e, \sin\alpha^s_e)$
        & Unit-circle projection of eye direction, avoiding angular discontinuity~\cite{hasan_mx-lstm_2018}.
        & Global & 2 \\

        \midrule

        \multirow{4}{*}{\shortstack[l]{Semantic\\targets}}
        & Gaze events
        & One-hot encoded label: attention, saccade, or noise.
        & -- & 3 \\
        & Presence of attention
        & Binary indicator: whether gaze is on any task-related object.
        & -- & 1 \\
        & Attention on traffic
        & Binary indicator: whether gaze is on shuttles (vs.\ other objects).
        & -- & 1 \\
        & Attention distribution
        & One-hot encoded label: leader shuttle, follower shuttle, goal, or environment.
        & -- & 4 \\

        \midrule

        \multirow{3}{*}{\shortstack[l]{Head\\orientation}}
        & Head-in-space $\alpha^s_h$
        & Absolute head direction angle in the world coordinate system.
        & Global & 1 \\
        & Head-in-walking $\alpha^w_h$
        & Head direction angle relative to the pedestrian's walking direction.
        & Local & 1 \\
        & Head vislet $(\cos\alpha^s_h, \sin\alpha^s_h)$
        & Unit-circle projection of head direction, analogous to eye vislet.
        & Global & 2 \\

        \bottomrule
    \end{tabularx}
\end{table*}

\section{Virtual Reality Experiment} \label{sec:experiment}

To test our model, we conducted an immersive VR experiment focusing on the specific pedestrian--automated shuttle interactions in shared space. We examined how various factors influence such interactions and collected data for later analysis.
Due to the lack of eye-tracking data in existing real-world pedestrian trajectory prediction datasets, we conduct a VR experiment to fulfill this purpose. 

\subsection{Experiment design} 

\subsubsection{Virtual environment}
\begin{figure*}[t]
    \centering
    \includegraphics[width=.85\linewidth]{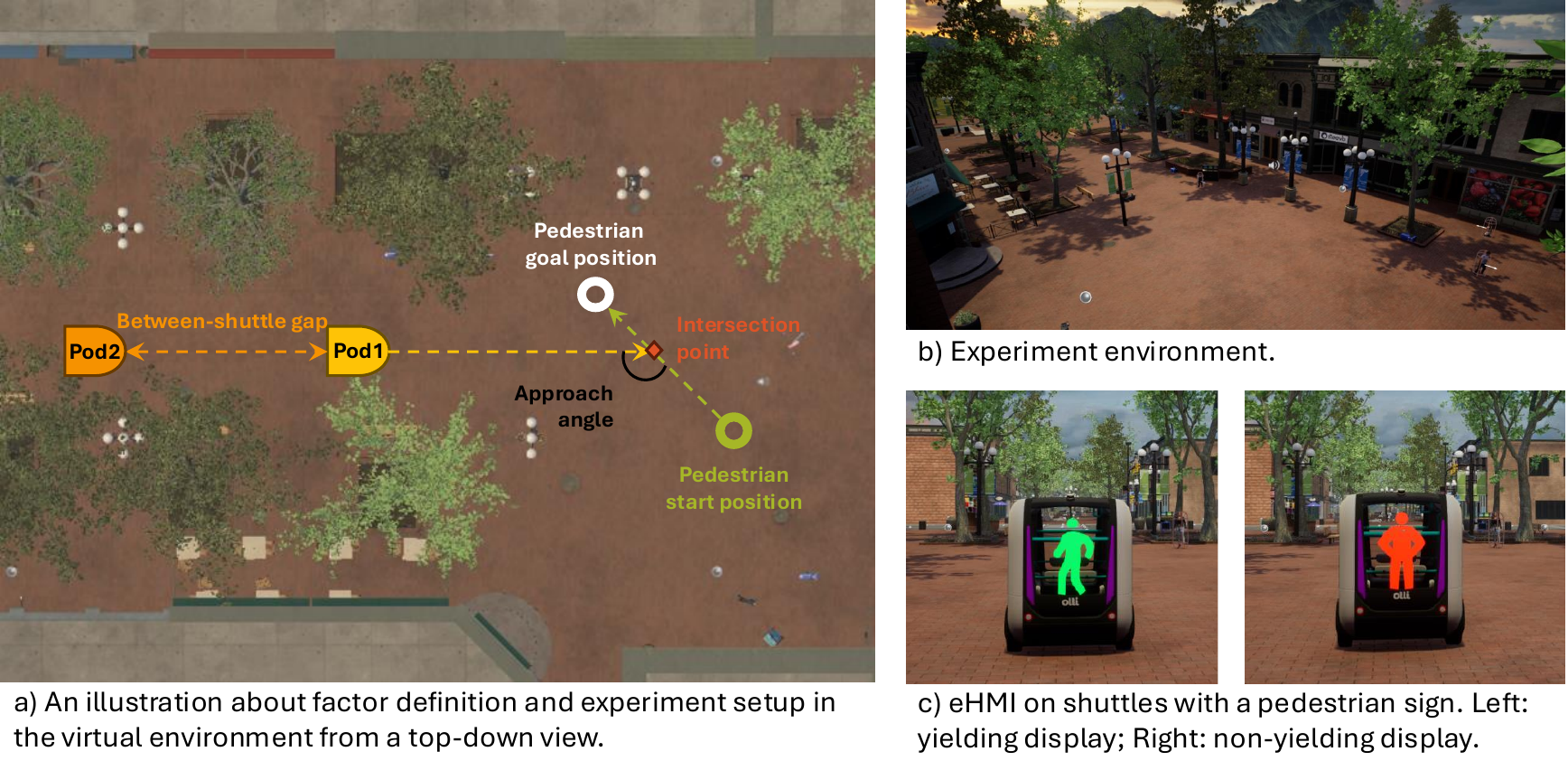}
    \caption{An overview of the experiment setup.}
    \label{fig:setup}
\end{figure*}

The virtual environment replicated a shared space resembling shopping streets in Delft, the Netherlands, built in Unreal Engine 5 (Fig.~\ref{fig:setup}b). The environment contained no right-of-way indicators: no traffic lights, stop signs, pedestrian crossings, or lane markings. Ambient audio simulating a busy shared space and electric shuttle sounds were added to enhance immersion.

\subsubsection{Factor design}
We manipulated four within-subject variables, as summarized below and illustrated in Fig.~\ref{fig:setup}a:

\paragraph{Shuttle behavior (2 levels)} 
This variable was selected because prior literature identifies vehicle yielding as a primary determinant of pedestrian crossing decisions \cite{anderson_off_2020}.
In our experiment, the shuttle either yielded or did not yield to the pedestrian. Non-yielding shuttles maintained a constant speed, whereas yielding shuttles followed a type~I deceleration profile~\cite{feng_does_2024}, initiating deceleration upon detecting the pedestrian at a distance of 12 meters.

\paragraph{eHMI presence (2 levels)} 
This factor was included because eHMIs have been shown to influence pedestrian crossing behavior \cite{kooijman2019ehmis, dey2020distance, feng2023effect}. 
When activated, the eHMI displayed a pedestrian symbol on the shuttle's front window (Fig.~\ref{fig:setup}c): a green signal indicated the shuttle would stop; a red signal indicated it would not. This design followed~\cite{feng_does_2024} and was chosen for its use of familiar, language-independent symbols~\cite{cumbal_crowdsourcing_2025}. The eHMI functionality was explained to participants during the familiarization phase.

\paragraph{Approach angle (3 levels)}
This factor is grounded in the eye-head-body coordination literature \cite{luo_eye-head-body_2022}, which shows that the angular relationship between a target and an individual’s current orientation influences visual attention allocation. 
Here, the angle between the shuttle's and pedestrian's forward directions was set to 45\textdegree, 90\textdegree, or 135\textdegree, following \cite{luo_eye-head-body_2022}. The 90\textdegree\ condition represents a standard perpendicular crossing. The 45\textdegree\ and 135\textdegree\ conditions were included to vary the visibility between pedestrian and shuttle and to elicit different eye-head-body coordination patterns.

\paragraph{Continuous traffic (3 levels)} 
This variable is motivated by gap-acceptance literature, which identifies the $3\sim7$~second interval as a critical range of ambiguity \cite{rasouli_autonomous_2020}.
In the experiment, participants encountered either a single shuttle or two consecutive shuttles separated by a temporal gap of 3 or 5 seconds. 
These two values bracket the critical range reported in prior pedestrian--AV gap-acceptance studies \cite{nunez_velasco_studying_2019, rasouli_autonomous_2020}. Our pilot study further confirmed that 3-second gap elicits predominantly cautious behavior while 5-second gap elicits mixed accept/reject decisions, making the pair well-suited to probe ambiguous conditions.


\subsubsection{Sampling}
The full factorial design yields 36 scenarios. Exposing each participant to all 36 would produce substantial fatigue and simulator sickness and introduce strong order and learning effects that would threaten the validity of within-subject comparisons. Based on pilot testing, we determined that 12 scenarios per participant was the maximum that could be completed with reduced confounds.

To preserve balance and ensure each factor level is adequately represented, we used \emph{Latin hypercube sampling} (LHS)~\cite{mckay_comparison_1979}, which stratifies each factor's range into equal intervals and ensures that every interval of every factor is sampled an equal number of times across the 12 trials per participant. This is standard practice in experimental design for high-dimensional factor spaces, and it yields more representative coverage than either random sampling or simple factorial reduction.

\subsubsection{Task}
Each trial began with a green circle (start) and a white circle (goal). Participants stepped into the start circle, oriented themselves toward the goal, and pressed the controller to begin. This initialization ensured that participants' eyes, head, and body were aligned toward the goal at trial onset, which is an important baseline for analyzing subsequent gaze deviations. Upon walking, the shuttle approached from 17.3~meters at 15~km/h, following the shared space speed limit in the Netherlands~\cite{zaken_participating_2024}. Participants were instructed to reach the goal safely, as they would in daily life. The trial ended upon reaching the goal. The next trial would be initiated. 

\subsection{Experiment apparatus}
The experiment was conducted in a $10 \times 5$~m room using an HTC VIVE Pro Eye headset (1440$\times$1600 pixels per eye, 110\textdegree\ field of view, 90~Hz refresh rate). Participants walked freely in the physical space with 1:1 mapping to the virtual environment (real-walking locomotion). The headset's built-in binocular eye tracker recorded gaze data at a precision of $0.5\sim1.1$\textdegree\; of visual angle.

\subsection{Experiment procedure}
\begin{figure*}[t]
    \centering
    \includegraphics[width=0.9\linewidth]{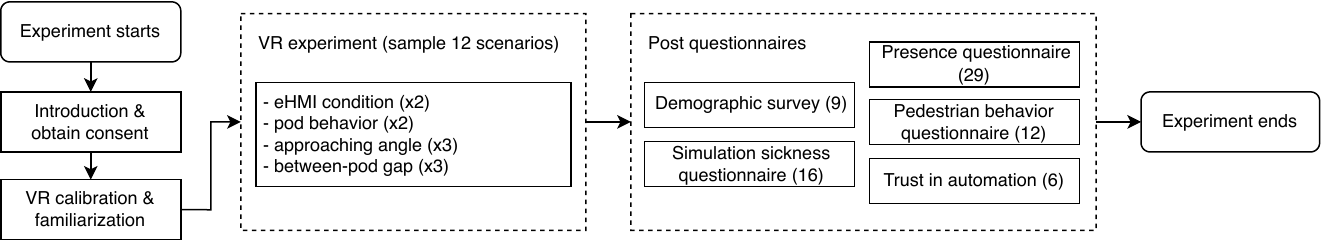}
    \caption{Experiment procedure. The numbers in the VR experiment block represent the levels of each variable, while the numbers in the post-questionnaire block indicate the number of items in each questionnaire.}
    \label{fig:procedure}
\end{figure*}

The procedure is illustrated in Fig.~\ref{fig:procedure}. Participants first received written information about the study and signed an informed consent form. After VR headset calibration, they completed a two-stage familiarization: (1) free exploration of the virtual environment, and (2) practice trials with the shuttle and eHMI. In the formal experiment, participants completed their 12 assigned scenarios in randomized order, returning to the start position between trials. Afterward, participants completed five questionnaires (Sec.~\ref{sec:data_collection}). Each participant received 15 euros as compensation. Ethical approval was granted by the Human Research Ethics Committee of Delft University of Technology (ID: 4888).

\label{sec:data_collection}
\subsection{Data collection}

\subsubsection{Objective data} 
Data recording began when participants left the start position and ended when they reached the goal. Data were sampled at 20~Hz and included: participant position, head orientation, eye gaze direction, fixation point, and fixation object.

\subsubsection{Subjective data} Five questionnaires were administered after the VR session: 
1) demographics (age, gender, nationality, education, handedness, walking habits, VR experience, familiarity with automated shuttles); 
2) the Simulation Sickness Questionnaire (SSQ) \cite{kennedy_simulator_1993} measuring experienced simulation sickness of participants in the virtual environment; 
3) the Presence Questionnaire (PQ) \cite{witmer_factor_2005} measuring the feeling of presence in the virtual environment;
4) the Pedestrian Behavior Questionnaire (PBQ) \cite{deb_development_2017} measuring violation, lapse, and positive behavior tendencies; and 
5) trust in automation \cite{payre_fully_2016}. 

\subsection{Participants}
Fifty-one participants (27 male, 24 female; aged 21--61, $M = 26.62$, $SD = 5.76$) were recruited. All had normal or corrected-to-normal vision and normal mobility. No participant withdrew due to motion sickness. Demographic details are reported in Tab.~\ref{tab:participants}. Each participant completed 12 trials, yielding 612 trials in total. After excluding trials with false triggers, missing eye-tracking signals, or lost VR tracking, 537 valid trials remained.

\section{Implementation details} \label{sec:implementation}

\subsection{Evaluation metrics and prediction modes}
We evaluate trajectory predictions using two standard metrics. The \textit{average displacement error} (ADE) measures the mean Euclidean distance between predicted and ground-truth positions across all future time steps:
\begin{equation}
    \text{ADE} = \frac{1}{N \times T_f} \sum_{n=1}^{N} \sum_{t=T_p+1}^{T_p+T_f} \| \vecy_{nt} - \hat{\vecy}_{nt} \|_2
\end{equation}
The \textit{final displacement error} (FDE) measures the prediction error at the last time step only:
\begin{equation}
    \text{FDE} = \frac{1}{N} \sum_{n=1}^{N} \| \vecy_{n,T_p+T_f} - \hat{\vecy}_{n,T_p+T_f} \|_2
\end{equation}
where $\hat{\vecy}_{nt} \in \real^2$ denotes the predicted position and $\vecy_{nt}$ the ground truth for sample $n$ at time step $t$.

Following~\cite{salzmann_trajectron_2020}, we report results under two prediction modes: (1)~\textit{deterministic}, using the most likely output of the predicted distribution, and (2)~\textit{stochastic}, using the best of 20 random samples from the predicted distribution (denoted as min$_{20}$). The deterministic mode reflects expected performance; the stochastic mode is used as a proxy for prediction uncertainty, where mean and standard deviation are reported over three random seeds. 

\subsection{Baseline}
To isolate the contribution of gaze and situational context, we construct a motion-only baseline with matched model capacity.
The baseline uses the same LSTM architecture as GazeX-LSTM but is restricted to motion features (past trajectory and velocity) and shuttle distance as input, excluding eye gaze and higher-level contextual variables.

This design enables a controlled comparison in which any performance differences can be directly attributed to the inclusion of gaze and contextual information, rather than architectural complexity. We adopt a lightweight LSTM for both models due to the moderate dataset size and the nature of our experimental setup, which involves a single pedestrian interacting with one or two shuttles. As such, the multi-agent social dynamics targeted by interaction-aware state-of-the-art models are intentionally absent in our setting.


\subsection{Data preparation}

\subsubsection{Motion data} \label{sec:motion_data_preprocessing}
Pedestrian trajectories and speed profiles are smoothed using a Gaussian filter ($\sigma = 4$). Trials in which the shuttle moved before the pedestrian (false triggers), or that contain missing eye-tracking or VR tracking signals, are excluded.

To segment the pre-crossing and crossing phases, we identify movement initiation points using a histogram-based velocity thresholding method~\cite{bindschadel_interaction_2021}. When multiple initiations occur within a trial, we use the last initiation before the pedestrian enters the shuttle's path as the phase boundary. The same procedure is applied to determine the backward-step velocity threshold.

\subsubsection{Eye gaze data}
Eye orientation data is preprocessed following recommendations in~\cite{grootjen_uncovering_2024}: missing values caused by blinks are filled via linear interpolation, and 0.1~s of data before and after each gap is removed to reduce blink-related artifacts. The head turning is determined using the I-VT method \cite{salvucci2000identifying} which thresholding the head movement speed. 

Semantic gaze targets are extracted from the raw fixation data obtained from the built-in eye trackers in three steps. First, saccades that involve the high-speed ballistic eye movement are identified using the I-VT method~\cite{salvucci2000identifying} with a threshold of 100\textdegree/s on angular eye-in-world velocity. Second, isolated fixation detections on non-saccade segments are smoothed by replacing them with the nearest semantic target. Third, gaps shorter than 0.4~s~\cite{schiffman1990sensation} flanked by identical targets are filled (likely blinks), while remaining segments shorter than 0.1~s are labeled as noise. This yields clean categorical labels indicating gaze allocation to the leader shuttle, follower shuttle, goal, or environment.

\subsubsection{Feature encoding}
Experimental variables (eHMI presence, shuttle yielding, approach angle, continuous traffic) are categorically encoded. Both the observation and prediction horizons are set to $T_p = T_f = 2$~s at 20~Hz, corresponding to 40 time steps each. This is consistent with horizons used in comparable pedestrian prediction work \cite{stein2022eye} and spans roughly $1/2\sim1/3$ of the typical pre-crossing interaction duration observed in our data, this giving the model access to the decision-relevant window.

\subsubsection{Data splitting}
The data is split by participant into training, validation, and test sets at a 6:1:3 ratio, ensuring that no participant appears in multiple splits. Within each trial, we extract samples using a sliding window with a stride of 4 for a trade-off between redundancy and information. This yields 11{,}916 training, 1{,}640 validation, and 3{,}918 test samples.

\subsection{Data analysis}

To characterize pedestrian behavior across experimental conditions, we use nine pedestrian behavioral metrics in five categories as shown in Tab.~\ref{tab:metrics}. 

\begin{table*}[t]
    \centering
    \caption{Metrics for characterizing pedestrian behavior in shared spaces.}
    \label{tab:metrics}
    \renewcommand{\arraystretch}{1.15}
    \begin{tabularx}{\textwidth}{
        >{\raggedright\arraybackslash}p{2cm}
        >{\raggedright\arraybackslash}p{3.5cm}
        >{\raggedright\arraybackslash}X
    }
        \toprule
        \textbf{Category} & \textbf{Metric} & \textbf{Description} \\
        \midrule
        
        \multirow{2}{2.5cm}{Movement} 
        & Gap selection & Gap chosen for crossing: before first shuttle, between shuttles, or after last shuttle. \\
        & Waiting time & Time from trial start to last movement initiation. \\
        \midrule
        
        \multirow{2}{2.5cm}{Hesitation} 
        & Initiation count & Number of crossing attempts, registered when velocity exceeds the initiation threshold (Sec.~\ref{sec:implementation}). \\
        & Backward count & Number of backward steps, registered when velocity falls below the backward threshold. \\
        \midrule
        
        \multirow{2}{2.5cm}{Deviation}
        & Mean deviation & Average lateral offset from the straight-line path connecting start and goal. \\
        & Max.\ deviation & Maximum lateral offset from the straight-line path. \\
        \midrule
        
        \multirow{2}{2.5cm}{Gaze}
        & Pre-crossing gaze time & Duration of shuttle-directed gaze before crossing initiation. \\
        & During-crossing gaze time & Duration of shuttle-directed gaze during crossing. \\
        \midrule
        
        Proxemics & Lateral clearance & Minimum lateral distance between participant and shuttle during passage. \\
        
        \bottomrule
    \end{tabularx}
\end{table*}

To assess the statistical significance of the observed gains in our models, we apply three paired one-tailed tests: (1) a paired bootstrap test \cite{efron1994introduction}, which resamples paired trial-level differences with replacement 10,000 times; (2) a Wilcoxon signed-rank test \cite{wilcoxon1992individual}, a non-parametric test on the signed ranks of paired differences; and (3) a permutation test \cite{good2005permutation}, which randomly permutes the pairing between the two models' per-trial errors 10,000 times to build a null distribution of the mean difference. Using three tests with different assumptions guards against any single test's sensitivities.

\subsection{Model interpretation}

To understand how situational variables influence predictions, we apply SHAP (SHapley Additive exPlanations)~\cite{lundberg2017unified}, which quantifies each feature's contribution to individual predictions using game-theoretic principles. A positive SHAP value indicates that the feature pushes the predicted position in the positive direction along the corresponding axis.

We use GradientExplainer to estimate Shapley values via expected gradients, with 100 random background samples. SHAP values are computed at the final predicted time step ($t = T_p + T_f$) under the deterministic prediction mode, as this time step corresponds to the position closest to the goal and captures the cumulative effect of the experimental variables. Results are visualized using 50 randomly selected test samples.
\section{Analysis and Results} \label{sec:results}

\subsection{Descriptive analysis of pedestrian behavior}
\label{sec:descriptive_analysis}

Descriptive statistics are summarized in Tab.~\ref{tab:descriptive_analysis}.
Shuttle yielding behavior generally facilitates crossing and reduces hesitation behavior, as indicated by shorter waiting times, fewer initiation and backward steps, smaller path deviations. All indicate that a shuttle's clear intent to stop lowers the decision burden. 
However, pedestrians in yielding conditions spend more time watching the shuttle during crossing and maintain larger lateral clearance. One plausible interpretation of this dissociation of reduced pre-crossing vigilance but increased crossing-phase monitoring is that yielding shifts the attentional profile from pre-crossing decision-making to in-crossing safety monitoring.

Non-standard angles (45\textdegree\ and 135\textdegree) consistently produce more hesitation, larger deviations, and reduced lateral clearance compared to the standard 90\textdegree\ crossing. Pedestrians appear to ``correct'' their path toward a perpendicular geometry, inflating measured deviations. The 135\textdegree\ condition is particularly instructive: it produces the largest path deviations and smallest lateral clearance despite offering the best natural visibility of the shuttle. We attribute this to an assertiveness effect --- when the shuttle is clearly visible early, pedestrians commit to crossing and accept tighter margins. Gaze patterns confirm this: pre-crossing gaze time is shorter at 135\textdegree\ (shuttle already in view) but longer during crossing (actively managing the tight clearance).


A 5-second gap between two shuttles generates more backward steps than either a single shuttle or a 3-second gap, despite feeling longer and more crossable. This difficulty pattern suggests the 5-second window is long enough to invite commitment but short enough to trigger second-guessing, which can be deemed as a characteristic of ambiguous affordance conditions in pedestrian behavior research and highlight the added complexity in decision-making under such conditions. 


The presence of eHMI has minimal observable effect across all behavioral metrics, with only a slight increase in backward steps. We return to this finding in the Discussion (Sec.~\ref{sec:discussion}).

In summary, these results show some behavioral variations in hesitation, spatial negotiation, and trajectory deviations across different experimental conditions, revealing empirical evidence for pedestrians' critical responses to ambiguous situations. 

\begin{table*}[t]
    \centering
    \caption{Descriptive statistics for movement, hesitation, deviation, gaze, and 
    proxemics variables. Values are reported as mean \stdb{SD} unless noted as counts. [] denotes the unit for each measure. 
    }
    \label{tab:descriptive_analysis}
    \footnotesize
    \setlength{\tabcolsep}{4pt}
    \renewcommand{\arraystretch}{1.15}
    \begin{tabular}{@{}l cc cc ccc ccc@{}}
        \toprule
        & \multicolumn{2}{c}{\textbf{Shuttle behavior}} 
        & \multicolumn{2}{c}{\textbf{eHMI}} 
        & \multicolumn{3}{c}{\textbf{Approach angle}} 
        & \multicolumn{3}{c}{\textbf{Continuous traffic}} \\
        \cmidrule(lr){2-3} \cmidrule(lr){4-5} \cmidrule(lr){6-8} \cmidrule(lr){9-11}
        \textbf{Measure}
        & Non-yield & Yield 
        & Absent & Present 
        & 45\textdegree & 90\textdegree & 135\textdegree 
        & 1 shuttle & 2 sh.(3\,s) & 2 sh.(5\,s) \\
        \midrule
        \multicolumn{11}{@{}l}{\textit{Movement behavior}} \\
        \quad Gap: before first shuttle [n]  
            & 82  & 261 & 178 & 165 & 117 & 106 & 120 & 101 & 121 & 121 \\
        \quad Gap: between shuttles [n]      
            & 109 & 4   & 54  & 59  & 25  & 51  & 37  & 79  & 2   & 32  \\
        \quad Gap: after shuttles [n]        
            & 80  & 1   & 37  & 44  & 31  & 26  & 24  & 0   & 55  & 26  \\
        \quad Waiting time [s]               
            & 6.7\,\stdb{4.3} & 4.3\,\stdb{3.3} & 5.5\,\stdb{4.1} & 5.5\,\stdb{4.0} 
            & 5.5\,\stdb{4.0} & 5.7\,\stdb{3.9} & 5.3\,\stdb{4.2} 
            & 5.0\,\stdb{3.2} & 6.0\,\stdb{4.4} & 5.5\,\stdb{4.3} \\
        \addlinespace\multicolumn{11}{@{}l}{\textit{Hesitation behavior}} \\
        \quad Initiation frequency [n]       
            & 1.6\,\stdb{0.7} & 1.5\,\stdb{0.7} & 1.6\,\stdb{0.7} & 1.6\,\stdb{0.7} 
            & 1.6\,\stdb{0.7} & 1.5\,\stdb{0.6} & 1.6\,\stdb{0.7} 
            & 1.5\,\stdb{0.6} & 1.6\,\stdb{0.7} & 1.6\,\stdb{0.8} \\
        \quad Backward frequency [n]         
            & 0.2\,\stdb{0.4} & 0.1\,\stdb{0.2} & 0.1\,\stdb{0.3} & 0.1\,\stdb{0.4} 
            & 0.1\,\stdb{0.4} & 0.1\,\stdb{0.3} & 0.1\,\stdb{0.4} 
            & 0.1\,\stdb{0.3} & 0.1\,\stdb{0.3} & 0.2\,\stdb{0.4} \\
        \addlinespace
        \multicolumn{11}{@{}l}{\textit{Deviation behavior}} \\
        \quad Mean deviation [cm]            
            & 22.7\,\stdb{23} & 20.7\,\stdb{20} & 21.9\,\stdb{22} & 21.5\,\stdb{21} 
            & 25.1\,\stdb{26} & 13.0\,\stdb{8}  & 27.3\,\stdb{24} 
            & 20.5\,\stdb{18} & 23.1\,\stdb{24} & 21.5\,\stdb{23} \\
        \quad Max.\ deviation [cm]           
            & 47.8\,\stdb{44} & 43.8\,\stdb{37} & 46.4\,\stdb{41} & 45.2\,\stdb{41} 
            & 50.1\,\stdb{46} & 28.2\,\stdb{16} & 59.6\,\stdb{46} 
            & 41.7\,\stdb{32} & 49.8\,\stdb{47} & 45.4\,\stdb{43} \\
        \addlinespace
        \multicolumn{11}{@{}l}{\textit{Gaze behavior}} \\
        \quad Pre-crossing gaze [s]          
            & 5.3\,\stdb{3.7} & 3.5\,\stdb{2.9} & 4.4\,\stdb{3.5} & 4.4\,\stdb{3.4} 
            & 4.1\,\stdb{3.2} & 4.6\,\stdb{3.4} & 4.5\,\stdb{3.7} 
            & 4.0\,\stdb{2.7} & 4.9\,\stdb{3.8} & 4.3\,\stdb{3.6} \\
        \quad During-crossing gaze [s]       
            & 3.3\,\stdb{1.4} & 4.0\,\stdb{2.0} & 3.6\,\stdb{1.8} & 3.7\,\stdb{1.7} 
            & 3.6\,\stdb{1.9} & 3.3\,\stdb{1.5} & 4.1\,\stdb{1.8} 
            & 3.4\,\stdb{1.4} & 3.8\,\stdb{1.8} & 3.8\,\stdb{2.0} \\
        \addlinespace
        \multicolumn{11}{@{}l}{\textit{Proxemics behavior}} \\
        \quad Lateral clearance [cm]         
            & 111.8\,\stdb{45} & 128.5\,\stdb{78} & 111.1\,\stdb{48} & 113.2\,\stdb{45} 
            & 109.1\,\stdb{34} & 125.1\,\stdb{55} & 98.9\,\stdb{37} 
            & 114.2\,\stdb{48} & 108.5\,\stdb{41} & 113.2\,\stdb{49} \\
        \bottomrule
    \end{tabular}
\end{table*}

\subsection{Approach angle modulates eye-head-body coordination}
\label{sec:mechanistic_look}

Before turning to predictive results, we characterize how eye, head, and body rotations vary with approach angle in our data. This analysis serves two purposes: it provides a behavioral foundation for interpreting the angle-dependent gains reported in Sec.~\ref{sec:predictive_results}, and it lets us compare our VR observations against the eye-head-body coordination patterns documented in real-world studies.

\begin{figure}[t]
    \centering
    \includegraphics[width=0.9\linewidth]{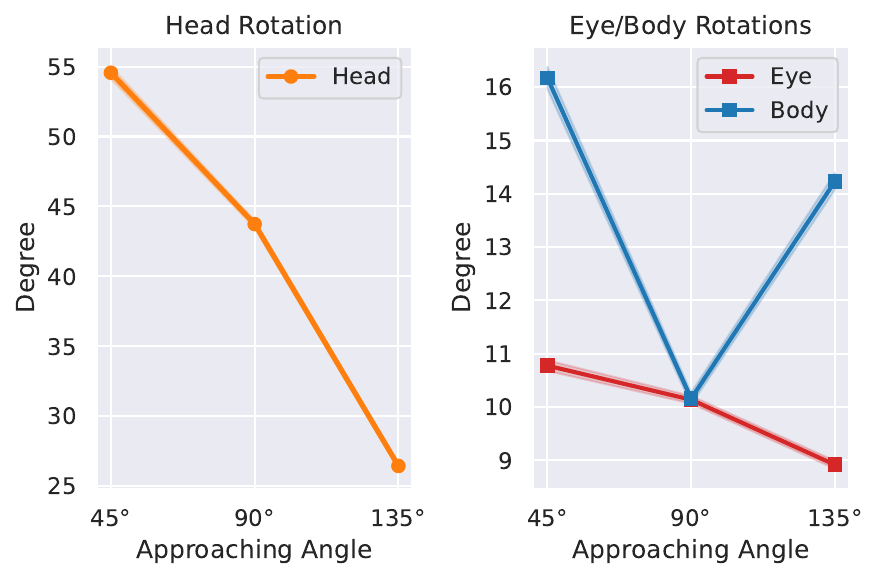}
    \caption{Analysis of mean rotation of eye, head, body under different approach angles.}
    \label{fig:eye_head_body_rotation_by_angles}
\end{figure}

Fig.~\ref{fig:eye_head_body_rotation_by_angles} shows mean rotation magnitudes for the three body segments across approach angles. Head rotation decreases sharply from 45° to 135°, consistent with the pattern documented in laboratory and real-world studies of eye-head coordination: as the target moves from the periphery toward the natural visual field, less head reorientation is required to foveate it.
Eye rotation follows the same decreasing trend but with smaller magnitude. We attribute this difference to VR's restricted field of view, which suppresses eye rotation relative to unconstrained environments \cite{postuma2025reduced}.

Body rotation shows a different pattern: rather than decreasing monotonically with angle, it dips at 90° and rises again at 135°. We interpret this as an artifact of the crossing task itself rather than a departure from the coordination literature. As the descriptive analysis showed (Sec.~\ref{sec:descriptive_analysis}), pedestrians at non-standard angles adapt their trajectories toward a more perpendicular crossing geometry, which increases body rotation relative to the direct path. At 90°, no such correction is needed; the natural crossing angle and the straight-line path align. The eye-head-body coordination documented in static tasks \cite{luo_eye-head-body_2022} thus combines with a goal-directed path correction in our locomotor task, producing the observed non-monotonicity for body rotation specifically.


\begin{figure}[t]
    \centering
    \includegraphics[width=0.9\linewidth]{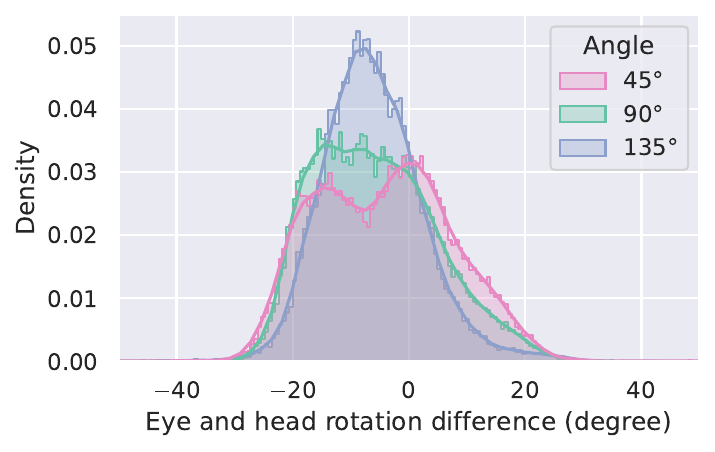}
    \caption{Distribution of relative eye–head angles under different approach geometries. Positive values indicate that gaze is directed further to the right than head orientation, whereas negative values mean that gaze is directed further to the left.}
    \label{fig:eye_head_rotation_diff_by_angles}
\end{figure}

Fig.~\ref{fig:eye_head_rotation_diff_by_angles} shows the distribution of eye-head angular differences at each approach angle.
Under 45\textdegree, the distribution is bimodal, consistent with participants either fully reorienting their heads along with their eyes or leaving a substantial eye-head offset.
Under 90\textdegree, the distribution is broader and flatter. Under 135\textdegree, it concentrates near zero where eyes and head are largely aligned, as expected when the target is within the natural visual field. Although the absolute eye-head divergence is smaller than in real-world studies \cite{luo_eye-head-body_2022}, reflecting the VR limitation noted above, the angle-dependent structure of the divergence is preserved.

\begin{figure*}[t]
    \centering
    \includegraphics[width=0.9\linewidth]{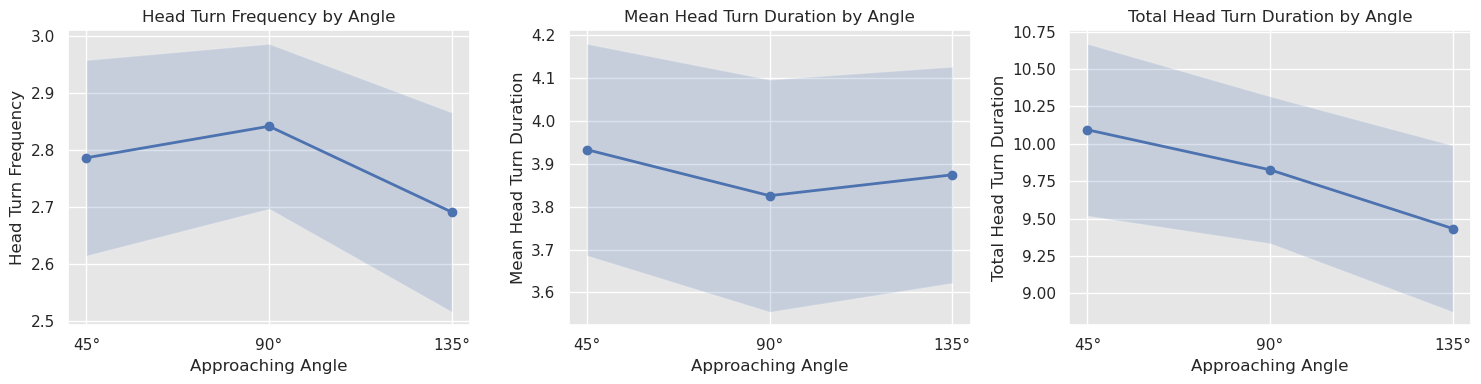}
    \caption{Head turn statistics under different approach angles.}
    \label{fig:head_turn}
\end{figure*}

Finally, Fig.~\ref{fig:head_turn} summarizes head-turning frequency and duration. At 45\textdegree, pedestrians produce fewer but longer head turns, consistent with sustained monitoring of a laterally approach shuttle. At 90\textdegree, turns become more frequent but shorter. At 135\textdegree, both frequency and total duration decline, since the shuttle is already in the visual field for much of the interaction. Total head-turning time decreases monotonically from 45\textdegree\; to 135\textdegree.


Taken together, these observations suggest that approach angle restructures how pedestrians allocate visual attention across their eye gaze, head, and whole-body rotations. Acute angles recruit the full eye-head-body chain; perpendicular angles rely more on the head and eyes; obtuse angles require minimal eye-head reorientation. This provides the behavioral basis for the predictive patterns reported next.


\subsection{Predictive modeling results}
\label{sec:predictive_results}

Tab.~\ref{tab:ablation_combined} summarizes all predictive results. Each row adds one information source to motion-only control. Percentage changes are relative to the baseline. These numbers achieve the same order of magnitude as recent real-world pedestrian prediction benchmarks when matched to our prediction horizon \cite{wang_ulptp-id_2026, ma_stochastic_2025}. The comparison is also evaluated on relative improvement over baselines within this controlled paradigm rather than to external benchmarks.

\begin{table*}[t]
    \centering
    \caption{Summary on the ablation study on the effects of eye gaze, head orientation, and situational context. All values are in centimeters. Numbers in bracket show the standard deviation. Percentage changes are relative to the baseline; \downpct{} indicates improvement, \uppct{} indicates degradation. The best result in each column is shown in \textbf{bold}.}
    \label{tab:ablation_combined}
    \renewcommand{\arraystretch}{1.2}
    \setlength{\tabcolsep}{0.55em}
    \begin{tabular}{@{}ll  ll  ll@{}}
        \toprule
         & & \multicolumn{2}{c}{\textbf{Deterministic prediction}} & \multicolumn{2}{c}{\textbf{Min$_{20}$ prediction}} \\
        \cmidrule(lr){3-4} \cmidrule(lr){5-6}
        \textbf{Category} & \textbf{Model configuration} & ADE $\downarrow$ & FDE $\downarrow$ & ADE $\downarrow$ & FDE $\downarrow$ \\
        \midrule

        ---
            & Baseline
            & 7.09 & 23.90
            & 3.31 \std{0.01} & 11.17 \std{0.12}\\

        \midrule
        \multirow{4}{*}{\shortstack[l]{Eye\\orientation}}
            & + Eye-in-space
                & 7.00 \downpct{1.19} & 23.32 \downpct{2.42}
                & 3.31 \std{0.01} \; \scriptsize{-0.0}     & 10.90 \std{0.10} \downpct{2.38} \\
            & + Eye-in-walking
                & 6.93 \downpct{2.18} & 23.28 \downpct{2.57}
                & 3.27 \std{0.02} \downpct{1.16} & 10.94 \std{0.08} \downpct{2.05} \\
            & + Eye vislet
                & 6.81 \downpct{3.84} & 22.95 \downpct{3.95}
                & 3.21 \std{0.00} \downpct{2.95} & 10.74 \std{0.07} \downpct{3.84} \\

        \midrule
        \multirow{4}{*}{\shortstack[l]{Semantic\\targets}}
            & + Gaze events
                & 7.14 \uppct{0.73}   & 24.13 \uppct{0.98}
                & 3.37 \std{0.02} \uppct{1.77}   & 11.34 \std{0.11} \uppct{1.53} \\
            & + Presence of attn.
                & 7.14 \uppct{0.73}   & 24.13 \uppct{0.98}
                & 3.37 \std{0.02} \uppct{1.77}   & 11.34 \std{0.11} \uppct{1.53} \\
            & + Attn.\ on traffic
                & 7.14 \uppct{0.74}   & 24.12 \uppct{0.93}
                & 3.34 \std{0.03} \uppct{0.76}   & 11.19 \std{0.06} \uppct{0.26} \\
            & + Attn.\ distribution
                & 6.98 \downpct{1.55} & 23.29 \downpct{2.56}
                & 3.28 \std{0.02} \downpct{1.06} & 10.94 \std{0.10} \downpct{2.06} \\

        \midrule
        \multirow{3}{*}{\shortstack[l]{Head\\orientation}}
            & + Head-in-space
                & 7.03 \downpct{0.75} & 23.78 \downpct{0.49}
                & 3.30 \std{0.02} \downpct{0.33} & 11.10 \std{0.07} \downpct{0.61} \\
            & + Head-in-walking
                & 6.91 \downpct{2.49} & 23.19 \downpct{2.96}
                & 3.27 \std{0.02} \downpct{1.23} & 10.89 \std{0.03} \downpct{2.50} \\
            & + Head vislet
                & 7.02 \downpct{0.99} & 23.69 \downpct{0.89}
                & 3.31 \std{0.02} \downpct{0.13} & 11.11 \std{0.05} \downpct{0.52} \\

        \midrule
        \multirow{1}{*}{\shortstack[l]{Situational\\context}}
            & + Context
            & 6.90 \downpct{2.56} & 22.66 \downpct{5.17}
            & 3.20 \std{0.02} \downpct{3.40} & 10.34 \std{0.03} \downpct{7.37} \\

        \midrule
        \multirow{3}{*}{\shortstack[l]{Combined}}
            & + Eye-in-space + context
                & 6.74 \downpct{4.83}          & 22.17 \downpct{7.24}
                & 3.16 \std{0.01} \downpct{4.57}          & 10.21 \std{0.13} \downpct{8.59} \\
            & + Eye-in-walking + context
                & \textbf{6.62} \downpct{6.53} & \textbf{21.88} \downpct{8.47}
                & \textbf{3.06} \std{0.02} \downpct{7.55} & \textbf{9.94} \std{0.08} \downpct{11.02} \\
            & + Eye vislet + context
                & 6.82 \downpct{3.71}          & 22.32 \downpct{6.61}
                & 3.21 \std{0.03} \downpct{2.93}          & 10.26 \std{0.13} \downpct{8.08} \\

        \bottomrule
    \end{tabular}
\end{table*}

\begin{table*}[t]
\centering
\caption{Paired significance tests comparing model pairs on per-trial FDE on deterministic predictions. 
Diff(A$-$B) is the mean difference in cm; negative values indicate that Model~A has lower error than Model~B. 
Bootstrap 95\% confidence intervals and $p$-values are computed from $10{,}000$ paired resamples over trials; 
Wilcoxon and permutation $p$-values are reported for robustness. 
All tests are one-sided ($H_1$: A~$<$~B).}
\label{tab:hypothesis_testing}
\setlength{\tabcolsep}{5pt}
\renewcommand{\arraystretch}{1.15}
\begin{tabular}{@{}llS[table-format=-1.4]cS[table-format=<1.4]S[table-format=<1.4]S[table-format=<1.4]c@{}}
\toprule
\textbf{Model A} & \textbf{Model B} & {\textbf{Diff (A$-$B)}} & \textbf{Boot 95\% CI} & {\textbf{Boot }\boldmath$p$} & {\textbf{Wilcox }\boldmath$p$} & {\textbf{Perm }\boldmath$p$} & \textbf{Sig.} \\
\midrule
+ Eye vislet             & Baseline            & -0.9449 & $[-1.19,\ -0.70]$ & <0.0001 & <0.0001 & <0.0001 & *** \\
+ Head-in-walking        & Baseline            & -0.7081 & $[-0.95,\ -0.47]$ & <0.0001 & <0.0001 & <0.0001 & *** \\
+ Eye vislet             & + Head-in-walking     & -0.2369 & $[-0.48,\ \phantom{-}0.00]$ &  0.0261 &  0.0092 &  0.0267 & *   \\
+ Eye-in-walking+context & + Eye vislet          & -1.0783 & $[-1.41,\ -0.75]$ & <0.0001 & <0.0001 & <0.0001 & *** \\
+ Eye-in-walking+context & + Context             & -0.7866 & $[-1.03,\ -0.55]$ & <0.0001 & <0.0001 & <0.0001 & *** \\
\bottomrule
\end{tabular}

\begin{tablenotes}[flushleft]
\footnotesize
\item Significance based on bootstrap $p$-values: * $p<0.05$, ** $p<0.01$, *** $p<0.001$. 
Bonferroni-corrected threshold for 5 tests per metric: $\alpha/5 = 0.010$; all comparisons marked *** remain significant after correction.
\end{tablenotes}
\end{table*}

\subsubsection{Eye gaze vs.\ head orientation} \label{sec:results_eye_vs_head}

Comparing the first three blocks below the baseline in Tab.~\ref{tab:ablation_combined}, we summarize our findings at below: 

\paragraph{Fine-grained gaze orientation improves prediction; semantic targets do not}
Overall, the incorporation of eye orientation improves behavioral prediction, although the effectiveness depends on how it is represented. 
Among the eye orientation representations, the eye vislet achieves the best performance (FDE: 22.95~cm vs.\ 23.32~cm baseline, $-3.95\%$), followed by eye-in-walking (FDE: 23.28~cm, $-2.57\%$). The absolute eye-in-space angle achieves less improvement over the baseline, suggesting that gaze direction is informative when encoded relative to the pedestrian's own movement or projected onto a continuous representation that avoids angular discontinuities. 

Among semantic gaze targets, only the finest-grained representation (attention distribution across four objects) yields marginal improvement. Coarser representations --- gaze events, presence of attention, and attention on traffic --- provide no benefit and even introduce noises, indicating that \textit{where} pedestrians look matters more for operational prediction than categorical labels of \textit{what} they look at.

\paragraph{Eye gaze provides predictive value beyond head orientation} \label{para:gaze_head_gain}
Head representation features alone significantly improve prediction over the baseline: the best head variant (head-in-walking) reduces FDE by 2.96\% to 23.19~cm. Eye gaze provides a further gain on top of this, with the best eye variant (eye vislet) reducing FDE by 3.95\%. The overall eye-vs-head difference on FDE is directionally consistent across all three tests (Wilcox $p=0.0092$, permutation $p=0.0267$, one-sided bootstrap $p=0.0261$; Tab.~\ref{tab:hypothesis_testing}), indicating that eye gaze contributes predictive information beyond what head orientation alone provides. A plausible mechanism is the temporal lead of the eyes over the head during gaze shifts, so eye-gaze features may encode pedestrian reorientation slightly earlier than head-based features. 

\paragraph{Eye-head gap varies with approach angle}
To further understand how eye gaze and head orientation help, Tab.~\ref{tab:decompose_metrics_by_angles} decomposes the predictive gains by approach angles. In general, both head and gaze information are more helpful on the acute angles (45\textdegree) compared to obtuse ones (135\textdegree), though this angle-dependent tendency is stronger when using eye gaze. Incorporating eye gaze information provides the largest improvement at 45\textdegree\ (FDE: $-5.36\%$), a moderate gain at 135\textdegree\ ($-3.80\%$), and the smallest at 90\textdegree\ ($-2.72\%$).
This might be explained by the eye-head-body coupling effect shown in Fig.~\ref{fig:eye_head_body_rotation_by_angles}: at acute angles, pedestrians must actively rotate their eyes (and eventually their heads) to monitor the shuttle, generating informative head and gaze dynamics about their crossing intentions. At obtuse angles, the shuttle falls within the natural visual field, thus producing subtler movements. 


Overall, this suggests that eye gaze signals improve the overall performance mostly by improving the acute angles prediction where the eye gaze is mostly activated and head-gaze difference is also the biggest. 

\begin{table*}[t]
    \centering
    \small
    \caption{Decomposing performance changes based on approach angles on the deterministic prediction. Comparison is between the baseline and 1) best model achieved by incorporating head info (head-in-walking), 2) best model achieved by incorporating eye gaze info (eye-vislet), 3) incorporating experimental info, 4) best model achieved by incorporating both eye gaze info (eye-in-walking) and experimental info.}
    \label{tab:decompose_metrics_by_angles}
    \setlength{\tabcolsep}{5pt}
    \renewcommand{\arraystretch}{1.15}
    \begin{tabular}{@{} l | l | llll @{}}
        \toprule
        \textbf{Metric} & \textbf{Type} & \textbf{45°} & \textbf{90°} & \textbf{135°} & \textbf{Overall} \\
        \midrule
        \multirow{2}{*}{ADE $\downarrow$} & Baseline & 8.39 & 6.23 & 6.92 & 7.09 \\
                             & + Head (head-in-walking) & 8.18 \downpct{3.09} & 6.10 \downpct{2.09} & 6.73 \downpct{2.83} & 6.91 \downpct{2.55} \\
                             & + Eye (eye vislet) & 7.94 \downpct{5.40} & 6.08 \downpct{2.30} & 6.66 \downpct{3.73} & 6.81 \downpct{3.95} \\
                             & + Context & 8.05 \downpct{4.09} & 5.99 \downpct{3.87} & 6.97 \uppct{0.75} & 6.90 \downpct{2.62} \\
                             & + Eye (eye-in-walking) + context & \textbf{7.83} \downpct{6.75} & \textbf{5.73} \downpct{7.96} & \textbf{6.57} \downpct{5.06} & \textbf{6.62} \downpct{6.59} \\
        \midrule
        \multirow{2}{*}{FDE $\downarrow$} & Baseline & 27.23 & 21.30 & 23.89 & 23.90 \\
                             & + Head (head-in-walking) & 26.39 \downpct{3.09} & 20.79 \downpct{2.37} & 23.08 \downpct{3.37} & 23.19 \downpct{2.97}\\
                             & + Eye gaze (eye vislet) & 25.77 \downpct{5.36} & 20.72 \downpct{2.72} & 22.98 \downpct{3.80} & 23.16 \downpct{3.97} \\
                             & + Context & 25.36 \downpct{6.85} & 19.93 \downpct{6.44} & 23.27 \downpct{2.58} & 22.66 \downpct{5.18} \\
                             & + Eye gaze (eye-in-walking) + context & \textbf{24.67} \downpct{9.42} & \textbf{19.31} \downpct{9.35} & \textbf{22.25} \downpct{6.85} & \textbf{21.88} \downpct{8.47} \\
        \bottomrule
    \end{tabular}
\end{table*}

\subsubsection{Effects of situational context} \label{sec:results_context}

Tab.~\ref{tab:ablation_combined} presents the effect of adding situational variables (approach angle, shuttle yielding, eHMI, continuous traffic) and Tab.~\ref{tab:effect_expt} shows how these gains vary with prediction horizon.
Situational context reduces FDE by $5.17\%$ overall (bootstrap $p<0.0001$, Wilcox $p<0.0001$, and permutation $p<0.0001$; Tab.~\ref{tab:hypothesis_testing}), but the improvement is concentrated at longer horizons: context slightly degrades accuracy at horizons below 1.0~seconds and reduces error more at 1.5~seconds and beyond. 
A possible explanation on this trade-off is: at short horizons, recent motion history is highly predictive on its own, and introducing additional static features may introduce noise. Conversely, at longer horizons, motion history becomes less informative, and scenario-level context becomes more valuable by bringing information about destination planning. 

The angle-decomposed analysis (Tab.~\ref{tab:decompose_metrics_by_angles}) further reveals that context helps most at 45\textdegree\ (FDE: $-6.85\%$) and 90\textdegree\ ($-6.44\%$) and least at 135\textdegree\ ($-2.58\%$). We hypothesize that at obtuse-angle encounters, the shuttle is clearly visible throughout the interaction and pedestrians behave more assertively. Acute and perpendicular angles where pedestrians face greater uncertainty benefit more from the explicit encoding of scenario constraints.

\begin{table}[t]
    \centering
    \caption{Effects of situational context variables on pedestrian behavior model over prediction horizons. `w/o' indicates the model without situational context, and `w/' indicates inclusion. Best result within each group is shown in \textbf{bold}.}
    \label{tab:effect_expt}
    \footnotesize
    \setlength{\tabcolsep}{4pt}
    \renewcommand{\arraystretch}{1.2}

    \subfloat[Deterministic prediction \label{tab:effect_expt_mean}]{%
    \begin{tabular}{@{}l cc cc cc cc@{}}
        \toprule
        & \multicolumn{2}{c}{\textbf{Baseline}} 
        & \multicolumn{2}{c}{\textbf{Eye-in-space}} 
        & \multicolumn{2}{c}{\textbf{Eye-in-walking}} 
        & \multicolumn{2}{c}{\textbf{Eye vislet}} \\
        \cmidrule(lr){2-3} \cmidrule(lr){4-5} \cmidrule(lr){6-7} \cmidrule(lr){8-9}
        \textbf{Horizon} & w/o & w/ & w/o & w/ & w/o & w/ & w/o & w/ \\
        \midrule
        0.5~s     & \textbf{0.27}  & 0.30           & \textbf{0.30} & \textbf{0.30}  & \textbf{0.27} & 0.31           & \textbf{0.29}  & 0.35           \\
        1.0~s     & \textbf{3.51}  & 3.56           & 3.63          & \textbf{3.48}  & 3.48          & \textbf{3.40}  & \textbf{3.38}  & 3.57           \\
        1.5~s     & 12.47          & \textbf{12.22} & 12.63         & \textbf{11.91} & 12.18         & \textbf{11.66} & \textbf{11.95} & 12.02          \\
        2.0~s & 23.90          & \textbf{22.66} & 24.43         & \textbf{22.17} & 23.28         & \textbf{21.88} & 22.95          & \textbf{22.32} \\
        ADE    & 7.09           & \textbf{6.90}  & 7.23          & \textbf{6.74}  & 6.93          & \textbf{6.62}  & \textbf{6.81}  & 6.82           \\
        \bottomrule
    \end{tabular}}

    \vspace{1em}

    \subfloat[Min$_{20}$ prediction \label{tab:effect_expt_min20}]{%
    \begin{tabular}{@{}l cc cc cc cc@{}}
        \toprule
        & \multicolumn{2}{c}{\textbf{Baseline}} 
        & \multicolumn{2}{c}{\textbf{Eye-in-space}} 
        & \multicolumn{2}{c}{\textbf{Eye-in-walking}} 
        & \multicolumn{2}{c}{\textbf{Eye vislet}} \\
        \cmidrule(lr){2-3} \cmidrule(lr){4-5} \cmidrule(lr){6-7} \cmidrule(lr){8-9}
        \textbf{Horizon} & w/o & w/ & w/o & w/ & w/o & w/ & w/o & w/ \\
        \midrule
        0.5~s     & \textbf{0.32} & 0.38           & 0.37          & \textbf{0.36}  & \textbf{0.32} & 0.38           & \textbf{0.36} & 0.43           \\
        1.0~s     & \textbf{2.24} & 2.26           & 2.30          & \textbf{2.25}  & 2.22          & \textbf{2.19}  & \textbf{2.19} & 2.30           \\
        1.5~s     & 5.35          & \textbf{5.16}  & 5.40          & \textbf{5.11}  & 5.30          & \textbf{4.92}  & \textbf{5.15} & 5.17           \\
        2.0~s & 11.17         & \textbf{10.34} & 11.35         & \textbf{10.21} & 10.94         & \textbf{9.94}  & 10.74         & \textbf{10.26} \\
        ADE    & 3.31          & \textbf{3.20}  & 3.38          & \textbf{3.16}  & 3.27          & \textbf{3.06}  & \textbf{3.21} & \textbf{3.21}  \\
        \bottomrule
    \end{tabular}}
\end{table}

\subsubsection{Complementarity of gaze and context} \label{sec:results_complementarity}

The most striking result emerges when eye gaze and situational context are combined (final row of Tab.~\ref{tab:decompose_metrics_by_angles}). The joint model (eye-in-walking + context) achieves an overall FDE reduction of $8.47\%$ over the baseline, larger than either source individually (eye gaze $3.95\%$, context $5.17\%$) and close to the sum of their separate contributions ($9.12\%$). This near-additive pattern indicates that the two signals capture largely distinct aspects of pedestrian behavior: situational context encodes the external constraints of the interaction, while eye gaze encodes the pedestrian's internal response to those constraints. This joint model is referred to as GazeX-LSTM in later use. 

\subsubsection{SHAP analysis on contextual variables}

\begin{figure}
    \centering
    \subfloat[Effect on x axis\label{fig:shap_analysis_x}]{%
       \includegraphics[width=\linewidth]{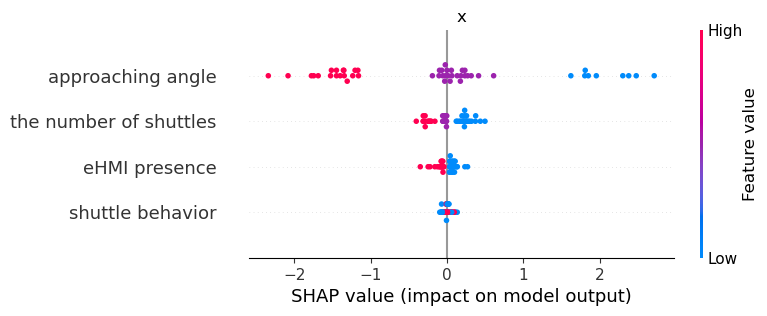}}
    \hfill

    \subfloat[Effect on y axis\label{fig:shap_analysis_y}]{%
       \includegraphics[width=\linewidth]{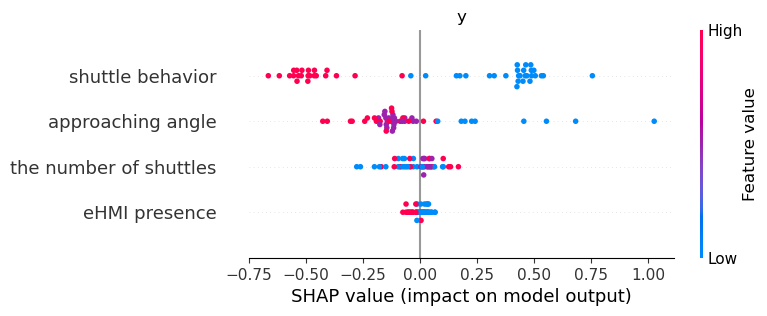}}
    \hfill
    \caption{SHAP analysis of the effect of experimental variables.}
    \label{fig:shap_analysis}
\end{figure}

To understand \textit{how} situational variables influence predictions, we apply SHAP analysis to the best-performing model. Fig.~\ref{fig:shap_analysis} shows SHAP values for the x-axis (horizontal movement toward or away from the shuttle) and y-axis (lateral movement indicating stepping forward or backward), respectively.

Approach angle is the dominant predictor on both axes. As expected, non-standard approach angles lead pedestrians to maintain the general crossing trajectory implied by the angle, but an interesting result is that pedestrians' avoidance strategies varied by angles: at 135\textdegree, they move assertively toward the shuttle (positive x-axis SHAP); at 90\textdegree, they shift slightly away horizontally; at 45\textdegree, they increase lateral distance. The 45\textdegree\ condition produces larger and more dispersed SHAP values than 135\textdegree, consistent with the greater behavioral uncertainty at acute angles observed in the descriptive analysis.

The number of shuttles has both lateral and longitudinal effects. With a single shuttle, pedestrians move away horizontally but approach laterally; with two shuttles, they consistently move backward regardless of the gap size. A wider gap additionally induce lateral approach, suggesting pedestrians incline to crossing between shuttles when the gap is sufficient.

Shuttle yielding produces a strong lateral effect: pedestrians reduce their lateral clearance when the shuttle yields, reflecting increased comfort. eHMI presence causes slight movement toward the shuttle on both axes, though its overall SHAP magnitude is the smallest among all variables, consistent with its weak effect in the descriptive analysis.

\subsubsection{Qualitative analysis}

\begin{figure*}[t]
    \centering
    \includegraphics[width=\linewidth]{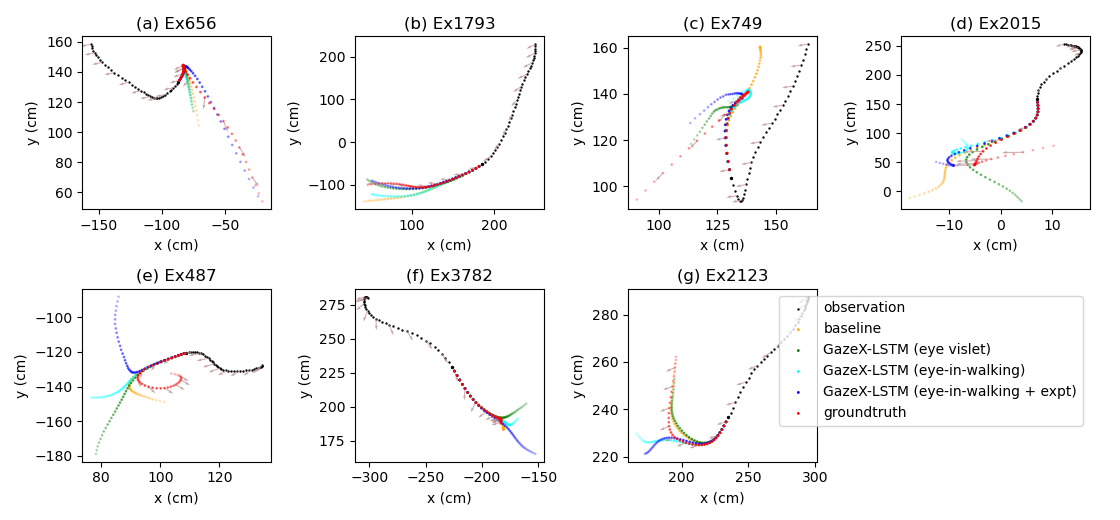}
    \caption{Qualitative comparison of predictions from \textcolor{Goldenrod}{baseline}, \textcolor{ForestGreen}{eye vislet variant} (best gaze-only representation), \textcolor{cyan}{eye-in-walking variant} (second-best gaze-only representation), and \textcolor{blue}{GazeX-LSTM} (eye-in-walking + context; overall best) against \textcolor{red}{ground truth}. \textcolor{brown}{Brown arrows} indicate pedestrian gaze direction from a top-down view; shuttles approach from the left side and moves in parallel to x-axis in all panels. The top row shows successful cases where adding gaze and context improves accuracy over the baseline; the bottom row shows failure cases.}
    \label{fig:vis}
\end{figure*}

Fig.~\ref{fig:vis} illustrates representative predictions. The top row (a--d) shows cases where incorporating eye gaze and context improves prediction: the model captures hesitation and subsequent recovery (a, c), maintains appropriate lateral spacing (b), and anticipates the onset of hesitation (d). The bottom row (e--g) shows failure cases where the augmented model performs worse than the baseline, typically involving sudden, unpredictable trajectory changes.


\subsection{Validity of the VR experiments}
To ensure the VR environment produced realistic behavior, we report four subjective measures. Results are summarized in Tab.~\ref{tab:subjective_measures}. These measures also serve to confirm psychological state of the sample cohort. 

\begin{table}[t]
    \centering
    \caption{Summary of subjective measures. SSQ subscales use weighted symptom scores; PQ subscales use a 7-point Likert scale (interface quality items are reverse-scored); trust uses a 7-point scale; PBQ uses a 5-point scale. Cronbach's $\alpha$ is reported for PBQ subscales.}
    \label{tab:subjective_measures}
    \renewcommand{\arraystretch}{1.15}
    \begin{tabular}{@{}llcc@{}}
        \toprule
        \textbf{Measure} & \textbf{Subscale} & \textbf{Mean} & \textbf{Standard deviation} \\
        \midrule

        \multirow{4}{*}{SSQ~\cite{kennedy_simulator_1993}}
            & Nausea          & 14.74  & 26.13 \\
            & Oculomotor      & 17.09  & 22.76 \\
            & Disorientation  & 35.48  & 35.55 \\
            & \textit{Total}  & \textit{164.54} & \textit{175.30} \\
        \midrule

        \multirow{5}{*}{PQ~\cite{witmer_factor_2005}}
            & Involvement        & 5.71 & 0.70 \\
            & Sensory fidelity   & 5.13 & 1.00 \\
            & Immersion          & 5.92 & 0.70 \\
            & Interface quality\textsuperscript{R} & 2.25 & 1.01 \\
            & \textit{Total}     & \textit{151.14} & \textit{17.57} \\
        \midrule

        Trust~\cite{payre_fully_2016}
            & ---             & 5.18 & 0.86 \\
        \midrule

        \multirow{3}{*}{PBQ~\cite{deb_development_2017}}
            & Violation \quad ($\alpha = 0.83$)  & 2.60 & 1.22 \\
            & Lapse \quad ($\alpha = 0.86$)      & 1.92 & 0.93 \\
            & Positive \quad ($\alpha = 0.73$)   & 4.22 & 1.02 \\

        \bottomrule
    \end{tabular}

    \vspace{0.3em}
    {\raggedright\footnotesize \textsuperscript{R}Reverse-scored items; lower values indicate better interface quality.\par}
\end{table}

\paragraph{Simulation sickness} The SSQ total score averaged 164.54 ($SD = 175.30$), with nausea receiving the lowest subscale score and disorientation the highest (likely due to repeated turning between trials). No participant reported notable discomfort.

\paragraph{Presence} The PQ total score averaged 151.14 ($SD = 17.57$), with immersion scoring highest ($M = 5.92$ on a 7-point scale), indicating strong engagement with the virtual environment.




\label{sec:trust_in_automation}
\paragraph{Trust in automation} Participants reported moderate trust in the automated shuttle ($M = 5.18$, $SD = 0.86$ on a 7-point scale), suggesting the shuttle was perceived as a realistic agent.


\paragraph{Pedestrian behavior} The PBQ showed good internal consistency for violation ($\alpha = 0.83$) and lapse ($\alpha = 0.86$) scales. Positive behavior consistency improved to acceptable levels ($\alpha = 0.73$) after removing one item. Mean scores (violation: 2.60, lapse: 1.92, positive: 4.22) indicate a generally cautious participant sample.


\section{Discussion} \label{sec:discussion}

\subsection{Eye gaze complements head orientation, with angle-dependent importance}

Consistent with prior work \cite{ridel_understanding_2019, herman_pedestrian_2022, hasan_mx-lstm_2018, mo2022multi}, head orientation alone improved prediction over the motion-only control in our data. Eye gaze provided an additional improvement on top of head orientation on overall FDE, though as noted in Sec.~\ref{para:gaze_head_gain}, this overall-level difference is directional rather than conclusive under multiple-comparison correction. 
This means eye gaze did not outperform head orientation everywhere significantly, but that the two signals carry distinguishable information under specific geometric conditions, which we explore in the following section.



The stronger form of the claim is angle-dependent. Our results show that eye gaze provides a larger predictive gain over head orientation at acute angles (45\textdegree), while the two signals perform comparably at obtuse angles (135\textdegree). This pattern is consistent with three related findings from the neuroscience and psychology literature. 
First, gaze shifts precede head rotation by $20\sim40~ms$ for horizontal movements of sufficient amplitude \cite{populin_target_2011, freedman2008coordination}. 
Second, the vestibulo-ocular reflex enables the eyes to reach and stabilize on a target faster than the head: once the eyes have fixated, the head is typically still catching up \cite{populin_target_2011}. 
Third, the eye-head difference is most pronounced when the target lies in the visual periphery, and diminishes when the target approaches the primary visual field \cite{luo_eye-head-body_2022}. 
Taken together, these evidence imply that head orientation and eye gaze carry overlapping information when the target is within the natural visual field (135\textdegree), and carry increasingly distinct information as the target moves toward the periphery. Our motion analysis in Fig.~\ref{fig:eye_head_body_rotation_by_angles} also confirms this pattern. 

Importantly, this eye-head gap we observe is likely a lower bound on what real-world deployment would reveal. VR's restricted field of view reduces eye-head divergence relative to natural environments \cite{postuma2025reduced, aizenman_statistics_2023}, so our participants likely under-rotated their eyes relative to what they would do in physical shared spaces. The advantage of eye-tracking-based signals over head-pose-only proxies should therefore be larger in real-world deployments than in our VR study.

\subsection{Continuous gaze orientation outperforms semantic fixation labels: a foveal-richness and peripheral vision explanation}

One of our main findings is that semantic gaze fixation labels (e.g., whether the pedestrian fixated the shuttle) underperformed continuous gaze orientation for trajectory prediction. AoI-based metrics remain the dominant approach in pedestrian gaze analysis and human-factors research \cite{lanzer_interaction_2023, bindschadel_two-step_2022}. However, our results suggest that this representation may be less suitable for motion prediction tasks. We propose two possible explanations, both related to how visual information supports human movement.

First, semantic fixation labels may discard information contained in continuous gaze direction. Both representations originate from the same signal: gaze orientation estimated from the pupil, which approximates the direction of foveal attention. The difference lies in how this information is encoded. 
AoI methods convert continuous gaze into binary assignments to predefined objects. In doing so, they remove information preserved in the continuous signal, including gaze trajectories between fixations, the direction and speed of gaze shifts, and cases where gaze is directed near an object without formally intersecting it. 
Prior work has shown that eye movements are closely linked to motor preparation, with gaze shifts often preceding head and body reorientation \cite{land2009looking, populin2011target}. These continuous dynamics may therefore contain information relevant for trajectory prediction that AoI labels do not capture. 
In addition, AoI extraction depends on threshold choices and object boundaries, which can introduce sensitivity to arbitrary design decisions \cite{hessels2017noise}.

Second, continuous gaze orientation may indirectly capture aspects of peripheral attention that semantic fixation labels cannot represent. Human vision is functionally divided into central and peripheral channels: foveal vision supports high-acuity object recognition, while peripheral vision is important for detecting motion and spatial changes \cite{strasburger2011peripheral}.
During navigation, humans rely on both channels simultaneously. Central vision is used to inspect task-relevant objects \cite{zhao2023pedestrian}, whereas peripheral vision supports monitoring of surrounding traffic and scene dynamics \cite{abbasi2021analysis}. 
Studies in driving research have shown that peripheral vision contributes to hazard detection and steering decisions even when attention is directed elsewhere \cite{crundall1999driving, wolfe2017more}.
Although eye trackers cannot directly measure peripheral perception, gaze direction and peripheral attention are closely related. For example, previous work has shown that incorporating near-peripheral field of vision through object-gaze distance can improve the estimation of visual attention \cite{wang2021object}.
We speculate that continuous gaze orientation may contain information correlated with peripheral monitoring that is lost when gaze is reduced to discrete fixation labels.

Together, these explanations suggest that AoI labels and continuous gaze orientation may serve different purposes.
AoI-based metrics are designed to answer whether a pedestrian explicitly attended to a particular object, which makes them useful for human-factors studies and interpretable behavioral analysis. 
In contrast, trajectory prediction depends on continuously estimating how pedestrians engage with the surrounding scene. For this purpose, continuous gaze orientation may provide a richer and more behaviorally relevant representation.

\subsection{eHMI effect and design implication}

Our eHMI produced limited effects on pedestrian behavior. Across most behavioral metrics, the eHMI-present and eHMI-absent conditions were statistically similar, with only a small increase in backward steps when eHMI was active (Tab.~\ref{tab:descriptive_analysis}). The SHAP analysis supports this finding: among the four situational variables, eHMI had the lowest feature importance on both spatial axes (Fig.~\ref{fig:shap_analysis}).

This result is consistent with part of the existing literature on eHMIs. Some studies report that eHMIs improve pedestrian crossing decisions, particularly when vehicle intent must be inferred at longer distances or before motion cues become clear \cite{kooijman2019ehmis, dey2020distance}. Other studies, however, suggest that eHMIs contribute little when vehicle kinematics already communicate intent effectively \cite{feng2023effect}. Two aspects of our experimental design may help explain the limited effect observed here.

First, the shuttles used a clear deceleration profile when yielding, making changes in speed and distance directly observable to pedestrians. In our case, the kinematic behavior alone may have been sufficient for participants to infer shuttle intent, leaving limited additional value for the eHMI. This interpretation is also consistent with the moderate-to-high trust ratings reported in Sec.~\ref{sec:trust_in_automation}, suggesting that participants were generally confident in interpreting shuttle behavior.

Second, participants encountered the eHMI repeatedly within the 12 trials. Although the familiarization phase ensured that participants understood the eHMI semantics, repeated exposure may also have reduced its salience over time. As a result, our findings likely reflect the effect of an already-familiar eHMI rather than the effect of a first encounter. 

In summary, our findings do not necessarily contradict studies reporting positive eHMI effects, but instead suggest possible boundary conditions for their effectiveness. In low-speed shared spaces where vehicle kinematics are easily observable and are at shorter distances, eHMIs may play a secondary role in communicating intent. 

\subsection{Limitations and future work}

We note three limitations of this work and one broader implication for future research.

\paragraph{Ecological validity and real-world transfer}
Our VR experiment required participants to walk between fixed start and goal locations. This enabled controlled testing of the experimental factors, but it may not capture the full complexity of real-world shared-space interactions. In addition, no public dataset currently provides synchronized fine-grained eye-tracking and trajectory data for pedestrian--AV interactions, preventing direct validation of whether our VR-derived findings transfer to on-road settings. Future work could address both limitations by combining more open-ended VR locomotion setups, such as omnidirectional treadmills, with real-world data collection and validation studies.

\paragraph{Sample and scenario composition}
Our 51 participants were primarily university students aged 21--35, resulting in a relatively homogeneous sample. Previous work suggests that age, culture, and mobility characteristics can influence attention allocation and crossing behavior \cite{wiczorek2016investigating, zito2015street, pecchini2015street, solmazer2020cross}. In addition, our experiment focused on a single pedestrian interacting with up to two shuttles, without social influence, group behavior, or peer coordination. Real shared spaces commonly involve such multi-agent interactions. Generalizing our findings will therefore require more diverse participant populations and more socially complex scenarios.

\paragraph{Signal modality} We focused on gaze direction and semantic fixation targets. Other eye-tracker outputs (pupil dynamics) and physiological measures (heart rate, skin conductance) could further enrich the human-perspective signal, especially in visually cluttered environments where any single modality may become unreliable \cite{kim2024gaitway}. 
Evaluating how these signals complement each other and how their usefulness changes with environmental complexity is a promising direction for future work.


\paragraph{Beyond perpendicular interactions} Finally, our results point to a more general gap in the field. Approach angle substantively restructures pedestrian behavior, both descriptively and through distinct eye-head-body coupling mechanistically. However, existing pedestrian-AV prediction literature and existing shared-space datasets remain dominated by perpendicular crossing scenarios \cite{golchoubian_pedestrian_2023, li2025analyzing}. 
Future datasets should therefore deliberately include non-perpendicular interaction geometries to better capture the behavioral diversity of unstructured shared spaces. Similarly, AV communication strategies may need to be evaluated separately across different interaction geometries.
\section{Conclusion} \label{sec:conclusion}

This paper investigated how fine-grained eye gaze data and situational context (i.e., approach geometry, vehicle behavior, eHMI presence, continuous traffic) jointly support pedestrian trajectory prediction on data collected for shared-space interactions with automated shuttles from VR. 
There are three main results. First, eye gaze and situational context are complementary sources of predictive information: context encodes what kind of interaction is unfolding while gaze reveals how the pedestrian is resolving it in real time. Second, the predictive gains yielded by incorporating eye gaze and head orientation into a motion-based model are angle-dependent, and largest where active visual acquisition is most required, a finding grounded in the eye-head-body coupling structure documented in our motion analysis; within this pattern, eye gaze cannot be fully substituted by head orientation at acute approach angles. Third, continuous gaze orientation substantially outperforms categorical semantic fixation labels.

Together, these results suggest advancing existing approaches with a human-centered perspective, as fine-grained attentional signals carry predictive information that trajectory data alone cannot recover. As Augmented Reality (AR)-based gaze sensing matures, integrating human perceptual state with situational context offers a path toward more adaptive and human-aware automated vehicle technologies.

\bibliographystyle{IEEEtran}
\bibliography{sections/reference}

\section*{Appendices}
\renewcommand{\thesubsection}{\Alph{subsection}}
\label{sec:appendix}

\subsection{Related work}

Related works on eye gaze in pedestrian behavior analysis are summarized in Tab.~\ref{tab:related_work}. 

\begin{table*}[htbp]
    \centering
    \caption{Related work on eye gaze in pedestrian behavior analysis. cVAE stands for conditional variational autoencoder, GAT for graph attention network, LMM for linear mixed model, and ANOVA for analysis of variance. Yaw angles show left and right movement and pitch angles for up and down movement.}
    \label{tab:related_work}
    \renewcommand{\arraystretch}{1.1} 
    \begin{tabularx}{\textwidth}{
        >{\RaggedRight\arraybackslash}p{1.5cm} 
        >{\RaggedRight\arraybackslash}p{0.7cm}
        >{\RaggedRight\arraybackslash}X 
        >{\RaggedRight\arraybackslash}p{1.4cm}
        >{\RaggedRight\arraybackslash}X
    }
        \toprule
        \textbf{Area} & \textbf{Paper} & \textbf{Gaze/Head Representations and Metrics} & \textbf{Model} & \textbf{Main Findings} \\
        \midrule
        
        \textbf{Pedestrian trajectory prediction} 
        & \cite{hasan_mx-lstm_2018} 
        & Head orientation is represented as vislets, defined as short sequences of head pose estimation. 
        & LSTM 
        & Head pose correlates strongly with movement at high speeds, although this correlation weakens when pedestrians slow down. \\
        \cmidrule(l){2-5}
        
        & \cite{ridel_understanding_2019} 
        & Head pose for pedestrian’s awareness using whether there is eye contact. 
        & LSTM 
        & Jointly using motion and head pose achieves lower prediction error. \\
        \cmidrule(l){2-5}
        
        & \cite{herman_pedestrian_2022}
        & Head and body poses 
        & LSTM, cVAE
        & Including head and body orientation significantly improves short-term prediction (1-2s), but not for long-term prediction (3-4s). \\
        \cmidrule(l){2-5}
        
        & \cite{mo2022multi} 
        & Relative yaw angle between two agents for interaction modeling and used as an edge attribute within its graph. 
        & GAT
        & Preserving these spatial relationships through edge attributes (including relative yaw) is vital when using exclusive coordinate systems. \\
        
        \midrule
        
        \textbf{Locomotion prediction} 
        & \cite{bremer2021predicting} 
        & Eye yaw and pitch angle, relative to current position. 
        & LSTM
        & Benefit of eye gaze data depended on the task and environment. Situations with changes in walking speed benefits from eye gaze data. 
        \\
        \cmidrule(l){2-5}

        & \cite{bremer_predicting_2024}
        & Eye yaw and pitch angles. Reference frame is set with forward axis being as the yaw orientation at the point of prediction. 
        & LSTM, Transformer 
        & Intrinsic data alone is enough for locomotion prediction. Eye-tracking provides better predictive cues than head orientation alone.\\ 
        \cmidrule(l){2-5}

        & \cite{stein2022eye}
        & The average head and body tracker yaw angle of the input sequence.
        & LSTM
        & Head synergies with gaze in certain scenarios. Both head and gaze orientation is effective predictor for locomotion. \\
        \cmidrule(l){2-5}
        
        & \cite{kim2024gaitway} 
        & Eye yaw and pitch angle, but the work did not specify the coordinate system. 
        & LSTM 
        & Gait-related data is enough for locomotion prediction. Gaze data increases errors under distracted cases and reduces in low distraction. \\
        
        \midrule
        
        \textbf{Pedestrian-AV interaction}
        & \cite{bindschadel_active_2022} 
        & Gaze for approximating mental workload: 1) Number of fixation; 2) Mean fixation duration; 3) Number of saccades.   
        & LMM
        & AV communication concepts significantly reduce pedestrians' mental workload. \\
        \cmidrule(l){2-5}
        
        & \cite{bindschadel_two-step_2022} 
        & Gaze for AV interface design: Proportion of fixation duration on AoIs.  
        & ANOVA 
        & Effective AV communication should follow the found visual focus pattern. \\
        \cmidrule(l){2-5}
        
        & \cite{lanzer_interaction_2023} 
        & Gaze for studying distraction effects: 1) Frequency of gaze at different AoIs; 2) Time spent looking at an AoI relative to the rest. 
        & ANOVA
        & Distracted pedestrians benefit from crossing group and do not benefit from eHMI. \\
        \cmidrule(l){2-5}       
        
        & \cite{tapiro_pedestrian_2020} 
        & Gaze for visual attention: Visual attention dispersion.
        & LMM
        & Visually loaded urban environments are risk factors for child pedestrians. \\ 

        \midrule

        \textbf{Ours} & & Systematically considered eye gaze orientation representation, and quantifies the effectiveness of head orientation as a proxy. & LSTM & Demonstrates that gaze-driven predictive gains are angle-dependent. Eye gaze provides additional value over head orientation, particularly at acute interaction angles. \\ 
        
        \bottomrule
    \end{tabularx}
\end{table*}

\subsection{Details of experiment setup}

\subsubsection{Calculation of shuttles' initial distance} 

\begin{figure}[h]
    \centering
    \includegraphics[width=\linewidth]{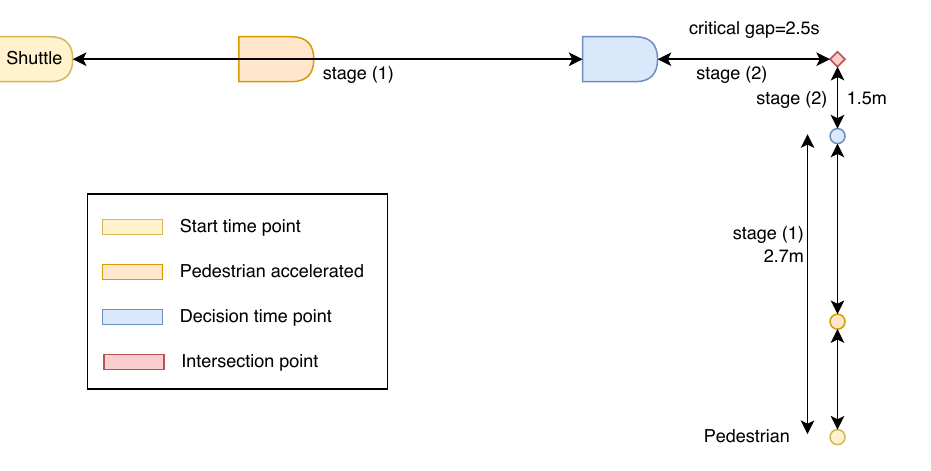}
    \caption{An illustration of the design of interactions.}
    \label{fig:init_dist_calulation}
\end{figure}

The process was split into two stages: (1) the pedestrian stood still, then accelerated, and maintained their speed until the decision point, and (2) the pedestrian decided whether to cross or not face the situation. So, for stage (1), it took the same time for pedestrians and shuttles to arrive at the blue point. For stage (2), the pedestrian had 1.5 meters to react, and the shuttle had a critical gap equal to 2.5 seconds. This resulted in a starting distance of the shuttle of 17.3 meters. The shuttle size is 3 meters $\times$1.6 meters.
Pedestrian average walking speed is 1.3 meters/second, and average acceleration is 1 meter/second \cite{zkebala2012pedestrian}. An illustration is shown in Fig.~\ref{fig:init_dist_calulation}. 

\subsubsection{Calculation of eHMI activation point} 

The eHMI was activated at a Time-to-Collision (TTC) of 2.5 seconds, a threshold widely recognized in traffic engineering for the activation of active safety systems \cite{lubbe_pedestrian_2014}. To translate it into distance, a red pedestrian eHMI sign appeared starting from 10.42 meters (=15km/h / 0.036 * 2.5 seconds); a green pedestrian eHMI sign appeared starting from 7.96 meters (as shown in the deceleration profile: when the shuttle was 7.96 meters away from the pedestrian, it needed 2.5 seconds to stop). The display persisted until the vehicle cleared the conflict point \cite{eisele2023equipping}.

\subsubsection{Sound design}
To ensure realism and immersion, there was audio in each scenario. Background ambient noise was used to simulate an outdoor shared space, and audio clips were used for the pod operating. 
The background sound is around 56 decibels \cite{misdariis_detectability_2013}. Vehicles approaching about 15 $km/h $ were detected at 27 meters for hybrids in electric mode \cite{kim_impact_2012} and reached about 55 decibels at 2 meters \cite{salleh_evaluation_2013}. To simplify the vehicle's sound effect implementation, we assumed the influence from distance and speed only. For distance, we adopted the sound level inversely proportional to the distance to the listener \cite{petiot_optimization_2019}. For speed, there is a log relationship between sound level and vehicle speed \cite{sandberg_are_2010}. 
The sound was displayed with spatial effects in the virtual environment, but it was not possible to distinguish sounds from more than two shuttles except by using distance cues. As Unreal Engine 5 was hard to achieve specific decibels, we conducted a pilot study to decide which sound level can realistically reflect the real-world sound effect.

\subsection{Participants' demographics information}

We summarize the participants' demographics in Tab.~\ref{tab:participants}. 

\begin{table}[t]
    \centering
    \caption{Demographic information of participants.}
    \label{tab:participants}
    \renewcommand{\arraystretch}{1.15}
    \setlength{\tabcolsep}{0.6em}
    \begin{tabular}{@{}llr@{}}
        \toprule
        \textbf{Characteristic} & \textbf{Category} & \textbf{Number of responses (\%)} \\
        \midrule

        \multirow{2}{*}{Gender}  
            & Male   & 27 (52.9) \\
            & Female & 24 (47.1) \\
        \midrule

        \multirow{2}{*}{Dominant hand} 
            & Right & 47 (92.2) \\ 
            & Left  &  4 \phantom{0}(7.8) \\ 
        \midrule
        
        \multirow{4}{*}{Education level}     
            & High school or equiv.    &  2 \phantom{0}(3.9) \\ 
            & Bachelor's or equiv.     & 19 (37.2) \\
            & Master's or equiv.       & 26 (51.0) \\
            & Doctoral or equiv.       &  4 \phantom{0}(7.9) \\
        \midrule

        \multirow{4}{*}{VR experience} 
            & Never     & 17 (33.3) \\
            & Seldom    & 23 (45.1) \\
            & Sometimes & 10 (19.6) \\
            & Often     &  1 \phantom{0}(2.0) \\
        \midrule
        
        \multirow{5}{*}{\shortstack[l]{Familiarity with\\automated shuttles}}
            & Not at all         &  5 \phantom{0}(9.8) \\
            & Slightly familiar  & 17 (33.3) \\
            & Moderately familiar & 19 (37.3) \\
            & Very familiar      &  7 (13.7) \\
            & Extremely familiar &  3 \phantom{0}(5.9) \\
        \midrule

        \multirow{4}{*}{\shortstack[l]{Experience with\\automated shuttles}}
            & None   & 27 (52.9) \\
            & Little & 13 (25.5) \\ 
            & Some   & 10 (19.6) \\ 
            & Extensive &  1 \phantom{0}(2.0) \\

        \bottomrule
    \end{tabular}
\end{table}

\subsection{Implementation details}

The hyperparameters of all models were tuned using the Optuna framework \cite{optuna_2019} on our dataset to make a fair comparison. All models were tuned over 150 trials for the best hyperparameters, and each trial runs 100 epochs. The learning rate decay is the same, which uses [15, 35, 60] as milestones, and the decay rate is 0.2. The optimizer is the Adam optimizer. The prediction is on the delta movement.

\end{document}